\documentclass{article}

\usepackage{microtype}
\usepackage{graphicx}
\usepackage{subfigure}
\usepackage{booktabs} 

\usepackage{tikz}
\usepackage{tabularx}
\usetikzlibrary {positioning}

\usepackage{url}

\usepackage{hyperref}
\usepackage{xcolor}

\newcolumntype{Y}{>{\centering\arraybackslash}X}

\usepackage[arxiv]{icml2024}

\usepackage{amsmath}
\usepackage{amssymb}
\usepackage{mathtools}
\usepackage{amsthm}

\usepackage[capitalize,noabbrev]{cleveref}

\theoremstyle{plain}
\newtheorem{theorem}{Theorem}[section]

\theoremstyle{definition}

\theoremstyle{remark}

\usepackage[textsize=tiny]{todonotes}

\icmltitlerunning{Weight-based Decomposition: A Case for Bilinear MLPs}

\begin{document}

\twocolumn[
\icmltitle{Weight-based Decomposition: A Case for Bilinear MLPs}




\icmlsetsymbol{equal}{*}

\begin{icmlauthorlist}
\icmlauthor{Michael T. Pearce}{indep}
\icmlauthor{Thomas Dooms}{indep}
\icmlauthor{Alice Rigg}{indep}
\end{icmlauthorlist}

\icmlaffiliation{indep}{Independent}

\icmlcorrespondingauthor{Michael Pearce}{michaelttpearce@gmail.com}

\icmlkeywords{Machine Learning, Mechanistic Interpretability, Spectral Decomposition, Language Models, Image Classification}

\vskip 0.3in
]



\printAffiliationsAndNotice{\icmlEqualContribution} 

\begin{abstract}
  Gated Linear Units (GLUs) have become a common building block in modern foundation models. Bilinear layers drop the non-linearity in the ``gate'' but still have comparable performance to other GLUs. An attractive quality of bilinear layers is that they can be fully expressed in terms of a third-order tensor and linear operations. Leveraging this, we develop a method to decompose the bilinear tensor into a set of sparsely interacting eigenvectors that show promising interpretability properties in preliminary experiments for shallow image classifiers (MNIST) and small language models (Tiny Stories). Since the decomposition is fully equivalent to the model's original computations, bilinear layers may be an interpretability-friendly architecture that helps connect features to the model weights. Application of our method may not be limited to pretrained bilinear models since we find that language models such as TinyLlama-1.1B can be finetuned into bilinear variants.
  
\end{abstract}

\section{Introduction} 
Multi-Layer Perceptrons (MLP) are ubiquitous components of large language models (LLMs) and deep learning models generally. Because of their nonlinear activation function, MLPs have historically been treated as irreducible components in mechanistic interpretability \cite{elhage_mathematical_2021, black_interpreting_2022}. 
Nonlinear activations are a major limitation because 1) they introduce complicated interactions (typically of all orders) between input features and 2) the neurons act as a privileged basis \cite{elhage_toy_2022, brown2023privileged}, so we cannot transform to a feature basis and cannot easily work with linear combinations of neurons. MLPs have only been well-understood in special cases such as small networks trained on math problems \cite{nanda2022progress, chughtai2023toy, michaud2024opening}. 

Recent interpretability work does not rely on parsing model weights. Instead, sparse autoencoders (SAEs) are used to derive interpretable features based on a model's internal activations over a dataset \cite{cunningham_sparse_2024, bricken_monosemanticity_2023, marks2024sparse}. Training transcoders to bridge features from one layer to those in the next ends up circumventing MLPs entirely \cite{dunefsky2024transcoders, sparsify}. Although SAE-based approaches have been successful, there remain questions about their data dependency and their out-of-distribution behavior. Ideally, there would be a way to connect the SAE-derived features to the model's computations.
 
Another approach is to use a more interpretable architecture. \citet{bilinear_note} suggested that bilinear MLP layers of the form $g(x) = (W\mathbf{x}) \odot (V\mathbf{x})$ have advantages for interpretability because their computations can be expressed in terms of linear operations with a third order tensor. This allows us to leverage tensor or matrix decompositions to understand the weights. A comparison of activations found that bilinear activations outperformed ReLU and GELU in transformer models \cite{shazeer2020glu}, and have performance only slightly below SwiGLU, which is prevalent in competitive open-source transformers today.

This paper aims to provide a demonstration of how bilinear MLP layers can be decomposed into a set of functionally relevant features. We use the word ``feature'' loosely to describe linear directions in activation space that appear interpretable, even if they are not necessarily sparse. Our contributions are as follows.\footnote{We share our code and other resources on \href{https://github.com/tdooms/bilinear-decomposition}{GitHub}}:

\begin{enumerate}
    \item We introduce a method to decompose bilinear MLPs into a set of eigenvector features with sparse interactions. This decomposition is fully equivalent to the layer's original computations. It is possible to chain these decompositions together to decompile a deep MLP-only model.
    \item In shallow image classification (MNIST) models, we show that the top eigenvectors are interpretable, and smaller eigenvalue terms can be truncated while preserving performance. Weight decay and noisy training produce qualitatively more interpretable features.
    \item Preliminary results on small language models (Tiny Stories) suggest that language eigenvectors can also be interpretable. We give evidence that larger pretrained language models can be finetuned into bilinear models using little data, to achieve loss similar to that expected from pretraining.
\end{enumerate}
We open-source our code and provide notebooks to replicate all experiments (we will provide a link here if accepted).

\section{Related Work} \label{sec:related}

\textbf{Weight-based Mechanistic Interpretability.} Reverse-engineering a model's inner workings from its weights has been a central goal of circuits-style research \cite{elhage_mathematical_2021, chughtai2023toy, olsson2022context, wang2022interpretability, michaud2024opening}. Our work is most similar in spirit to \citet{cammarata2020curve, cammarata2021curve}, which extracted simple curve features in CNNs from model weights and described how high-order features are constructed from simpler ones. Our work shares this basic approach but for an MLP-only model.  

\textbf{Transcoders.} Recently, transcoders have been used to derive sparse feature interactions by predicting an MLP layer's outputs from its inputs, effectively learning how to skip over the layer \cite{dunefsky2024transcoders, sparsify}. Our work has a similar goal of describing an MLP layer's outputs in terms of a sparse set of interactions, but we derive the interactions and input features directly from the weights, although our features do not necessarily have sparse activations. 

\textbf{Matrix Decompositions.} Various matrix decompositions have been used previously to understand neural network internals. In deep linear models, singular value decomposition (SVD) provides a nearly full description of how hierarchical semantic concepts are learned \citep{saxe2019mathematical}. Non-negative matrix factorization has been used to cluster neuron activations effectively \citep{olah2018the}. For transformers, it's been noticed that SVD of model weights often yields interpretable feature directions \citep{beren2022singular}. More recently, SVD has been used to decompose gradients and identify a sparse interaction basis \citep{bushnaq2024local, bushnaq2024using}.

\section{Background} \label{sec:bilinear}

We use a standardized notation as presented by \citet{tensor_decomp} throughout this paper. Scalars are denoted by a normal character $s$, vectors are denoted in bold $\textbf{v}$, matrices as capital letters $M$, and third-order tensors as calligraphic script $\mathcal{T}$. The entry in row $i$ and column $j$ of a matrix $M$ is a scalar and therefore denoted as $m_{ij}$. As is common in many computing libraries, we denote taking row $i$ or column $j$ of a matrix by $\textbf{m}_{i:}$ and $\textbf{m}_{:j}$ respectively. Additionally, for brevity, we use the term tensor to encompass all orders, wherever possible, we disambiguate. Lastly, we use $\odot$ to denote an element-wise product and $\cdot_\text{axis}$ to denote a product of tensors along the specified axis.
\subsection{Gated Linear Units} \label{background_glu}
Gated Linear Units (GLU), introduced in \citet{dauphin2017language}, use three weight matrices, as opposed to two for an ordinary MLP.
The output of an MLP layer using a GLU has the form $W_\text{out}g(\textbf{x})$, such that
\begin{equation} \label{eq:glu}
    g(\textbf{x}) = (W\textbf{x}) \odot \sigma(V\textbf{x}),
\end{equation}
where $\sigma$ is an activation function applied pointwise. Common
choices for $\sigma$ include Gaussian Error Linear Units, $\text{GELU}(x)=x\Phi(x)$,
and $\text{Swish}_\beta(x) = x \text{sigmoid}(\beta x)$.
Bilinear layers are the result of omitting the nonlinear activations $\sigma$. 

We have shown the activations without biases because the biases can be incorporated into the weights as an additional column: $W' = [W; b_w]$ and $x' = [x, 1]$. For ease of presentation, we consider models trained without biases, but our analysis methods can incorporate biases with extra bookkeeping.

\subsection{Interaction Matrices} \label{sec:background_tensor} 
To illustrate the behavior of bilinear layers, we show how a single neuron activation $g(\textbf{x})_a$ is computed.
\begin{align} \label{eq:interaction}
g(\textbf{x})       &= (W \textbf{x}) \odot (V \textbf{x}) \nonumber \\
g(\textbf{x})_a      
           &= (\textbf{w}_{a:}^T\textbf{x}) \ (\textbf{v}^T_{a:}\textbf{x}) \nonumber \\
           &= \textbf{x}^T \left(\textbf{w}_{a:} \textbf{v}_{a:}^T \right) \textbf{x} \nonumber
\end{align}

We call the term $B_{a::} = \textbf{w}_{a:} \textbf{v}_{a:}^T$
an \emph{interaction} matrix. This matrix defines how each pair of inputs interact for a given output. There is an interaction matrix for each output neuron, resulting in a third-order tensor $\mathcal{B}_\text{out,in,in}$ given by $b_{aij} = w_{ai} v_{aj}.$

\subsection{Symmetry} \label{sec:symmetry} 

\begin{figure*}
\centering
\includegraphics[width=0.9\textwidth]{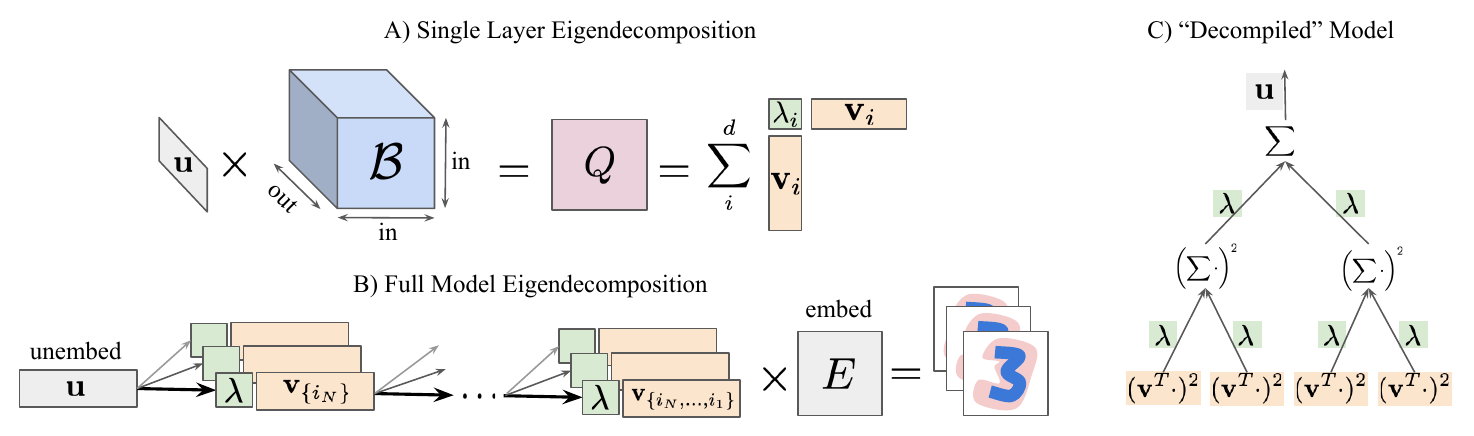}
\caption{Illustration of the eigenvalue decomposition. \textbf{A)} For a single layer, we start with a vector $\mathbf{u}$ in the output space of the third-order tensor $\mathcal{B}$. Their dot product gives a symmetric interaction matrix $Q$. Eigendecomposition of $Q$ gives an orthonormal set of eigenvectors $\mathbf{v}_i$. In this basis, the interactions are sparse since an eigenvector only interacts with itself. \textbf{B)} For a fully bilinear model, we start with an unembed vector $\mathbf{u}$ (e.g., for the digit ``3'') and repeatedly apply the single-layer eigendecomposition. Each layer $k$ eigenvector $\mathbf{v}_{\{i_N, \dots, i_k\}}$ acts as the output vector that determines a set of layer $k$-1 eigenvectors $\mathbf{v}_{\{i_N, \dots, i_k, i_{k-1}\}}$. Using the embedding weights, the layer-1 eigenvectors can be transformed into input features. \textbf{C)} Schematic of a 2-layer model after full decomposition showing the tree-like computational graph, with a branching factor of 2 (instead of d\_model) for simplicity. The model is \emph{decompiled} in the sense that interactions have been sparsified and made explicit through the graph. Only the layer-1 eigenvectors are needed to get the initial activations from the post-embedding inputs. The eigenvalue magnitudes parameterize the importance of the edges.}
\label{fig:eigendecomposition}
\end{figure*}

Since $g(\textbf{x})_a = \textbf{x}^T B_{a::} \textbf{x}$ is a scalar, we can write
\begin{align*}
    g(\textbf{x})_a
    &= g(\textbf{x})_a^T \\
    &= \textbf{x}^T B_{a::}^T \textbf{x}.
\end{align*}
Consequently, only the symmetric part of the interaction matrix, defined as $B_{a::}^{sym} = \frac{1}{2}(B_{a::}+B_{a::}^T)$, contributes to the output of $g(\textbf{x})_a$. 
The symmetric form will be useful because it has a nice eigenvalue decomposition. From here on, any interaction matrix is assumed to be in symmetric form, dropping the superscript. 

\begin{theorem}[Spectral Theorem]
If \(Q: \mathbb{R}^d \to \mathbb{R}^d\) is a real, symmetric matrix, then there exists an orthonormal basis of \(\mathbb{R}^d\) consisting of eigenvectors of \(Q\). Each eigenvalue is real. That is, \(Q=P^T\Lambda P\), where $\Lambda$ is a real diagonal matrix, and $P$ is a real orthogonal matrix.
\end{theorem}


\section{Analysis Techniques} \label{sec:spectral}

\subsection{Single-Layer Eigendecomposition} \label{sec:single_eigendecomp}

Directly studying the third-order tensor $\mathcal{B}$ is possible through higher-order tensor decompositions \cite{hosvd, tensor_decomp}, which we describe in \autoref{sec:tensor_decomp}. 

Higher-order techniques typically rely on first flattening the tensor into a matrix. Instead of flattening, we can reduce $\mathcal{B}$ using a vector $\mathbf{u}$ in the output space of $\mathcal{B}$. This yields a matrix $Q = \mathbf{u} \cdot_\text{out} \mathcal{B}$ that describes the interactions between input pairs and determines the output along the $\mathbf{u}$-direction. An advantage of reduction over flattening is that we can use meaningful $\mathbf{u}$ vectors, such as unembedding vectors (eg for specific MNIST digits) or downstream feature vectors.

As illustrated in \autoref{fig:eigendecomposition}, we can diagonalize the interaction matrix by taking its eigenvalue decomposition. Since $Q$ can be taken as symmetric (see \autoref{sec:symmetry}), the spectral theorem provides a decomposition of the form
\begin{align}
    Q = \sum_i^d \lambda_i \ \mathbf{v}_i \mathbf{v}_i^T
\end{align}
with a set of $d$ (model dim) orthonormal eigenvectors $\mathbf{v}_i$ and real-valued eigenvalues $\lambda_i$.

Since an eigenvector only interacts with itself, the eigenvectors form a basis with sparse interactions, and the eigenvalues $\lambda_i$ parameterize the interaction strengths. In this basis, the output in the $\mathbf{u}$-direction is 
\begin{align}
    \mathbf{x}^T Q \mathbf{x} = \sum_i^d \underbrace{\lambda_i \ (\mathbf{v}_i^T \mathbf{x})^2}_{\text{activation for }\mathbf{v}_i}
\end{align}
where each term can be considered the activation for the eigenvector $\mathbf{v}_i$.

If we have a set of $\mathbf{u}$-vectors, it's possible to re-express $\mathcal{B}$ in terms of the eigenvectors, which amounts to a change of basis, as described in \autoref{sec:rewrite_b_tensor}. 

\subsection{Full Model Eigendecomposition}\label{sec:full_model_eigendecomp}

\begin{figure*}[t]
\centering
\includegraphics[width=0.95\textwidth]{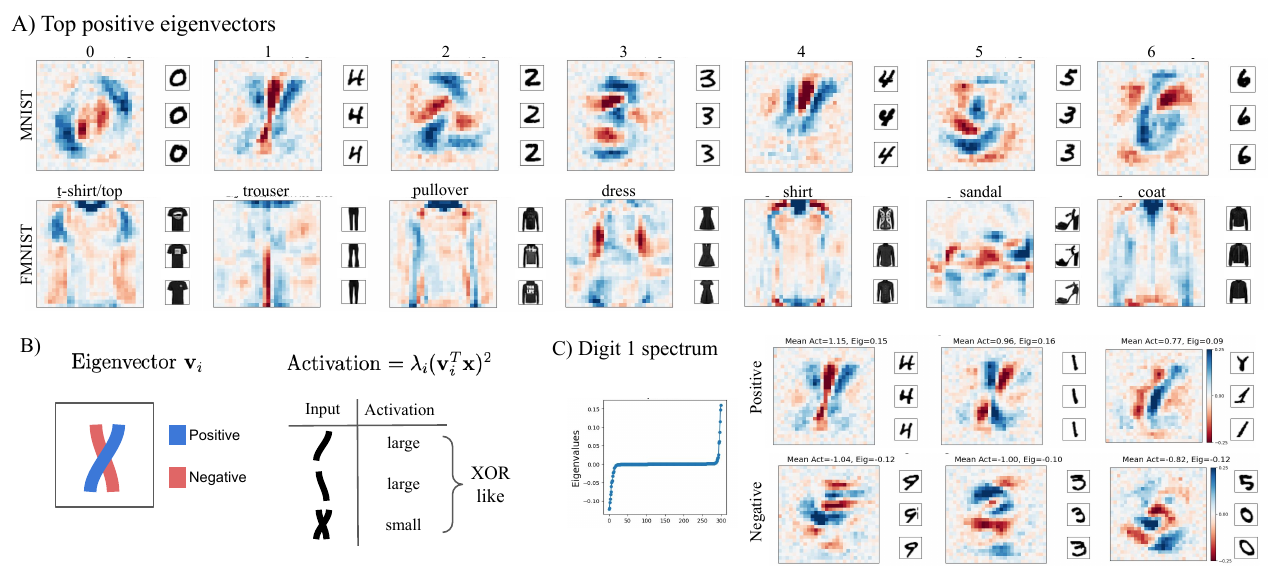}
\caption{Eigenvectors for single-layer MNIST and Fashion-MNIST models. \textbf{A)} Top positive eigenvectors by mean activation for the first seven output classes. To the right of each eigenvector are the 3 inputs with the highest activations. Ordering by activation favors eigenvectors with more ``complete'' images that overlap more with the inputs. Often the top activating eigenvector has the top or second top eigenvalue. \textbf{B)} Because of the square dot product in an eigenvector's activation, $\lambda_i (\mathbf{v}_i^T \mathbf{x})^2$, its overall sign is arbitrary, so we choose the sign so that the positive component (blue) is similar to the top inputs. If an input separately has a high overlap with the positive (blue) or negative (red) component then the activation can be large, but a high overlap with both components cancels out. This behavior is similar to an XOR. \textbf{C)} The top three positive and negative eigenvectors for digit 1. The top positive eigenvalue has strong activations for digit 4, but these likely cancel out with activations for the top negative eigenvalue which detects horizontal lines in the center of an image.}
\label{fig:mnist_eigenvectors}
\end{figure*}

\autoref{fig:eigendecomposition} (B) shows how a deep MLP-only model with bilinear layers can be decomposed by starting with the unembedding vectors $\mathbf{u}$ and repeatedly applying the single-layer eigendecomposition, moving in the direction of outputs to inputs. Each layer $k$ eigenvector $\mathbf{v}_{\{i_N, \dots, i_k\}}$ acts as the output vector that determines a set of layer $k$-1 eigenvectors $\mathbf{v}_{\{i_N, \dots, i_k, i_{k-1}\}}$. Finally, the layer 1 eigenvectors can be transformed into input features using the embedding weights and then interpreted. 

The decomposition process produces a tree-like computational graph for each starting unembedding vector, shown schematically in \autoref{fig:eigendecomposition} (C). Only the input features (layer-1 eigenvectors) are needed to get initial activations from the post-embedding inputs. We call this model \emph{decompiled} because the interactions between input features are explicitly captured in the graph, for example, two input features interact only in the layer where their branches join. The eigenvalue magnitudes parameterize the importance of edges and would allow the most important subgraphs to be identified. 

By itself, this type of decompilation is unlikely to be useful for realistic models with attention and very large models since the number of layer-1 eigenvectors grows exponentially with the number of layers. However, it may serve as a useful theoretical tool for studying higher-order feature construction and identifying submodules in MLP-only models.

\section{Image Classification} \label{sec:mnist}
We demonstrate our approach on the MNIST dataset of handwritten digits and the Fashion-MNIST dataset of clothing images. We use a model consisting of one or two bilinear layers with linear embedding and unembedding layers. Unless specified otherwise, the model dimension is 300. Early experiments showed that biases and normalization layers can be incorporated into the decomposition with similar results, but for simplicity, we have not included them in the models used for the results presented here. Details of the training setup can be found in \autoref{app:mnist_setup}. 

\subsection{Single-Layer Eigenvectors}
\autoref{fig:mnist_eigenvectors} shows the top positive eigenvectors for the output classes in each dataset. Since the activation of an eigenvector, $\lambda_i (\mathbf{v}_i^T \mathbf{x})^2$, contains a squared dot product, both positive and negative components can contribute to large activations. Cancellations between them lead to small activations. So the form of the activation gives an XOR-like structure, as illustrated in (B). The squared dot product ensures that the activations for positive eigenvectors are always positive, and those for negative eigenvectors are always negative.

Many of the MNIST eigenvectors resemble the full digit (0,2,3,6) while others emphasize certain parts (4,5). The FMNIST eigenvectors focus much more on specific parts, for example, the neck and sleeve areas on shirts/pullovers/coats and the leg gap for trousers.

A portion of the eigenvalue spectrum is shown for digit 1 in \autoref{fig:mnist_eigenvectors}(C). Only a small fraction of eigenvalues have significant magnitude. For the first positive eigenvector, the top three inputs by activation are all 4s. The first negative eigenvector activates on horizontal lines in the middle, so we expect it to cancel out the activations on 4s for the first positive eigenvector. The other positive eigenvectors contain 1-like features at different slants.  Similar plots for the other digits can be found in \autoref{app:mnist_eigs}. 

\begin{figure}[t]
\centering
\includegraphics[width=0.35\textwidth]{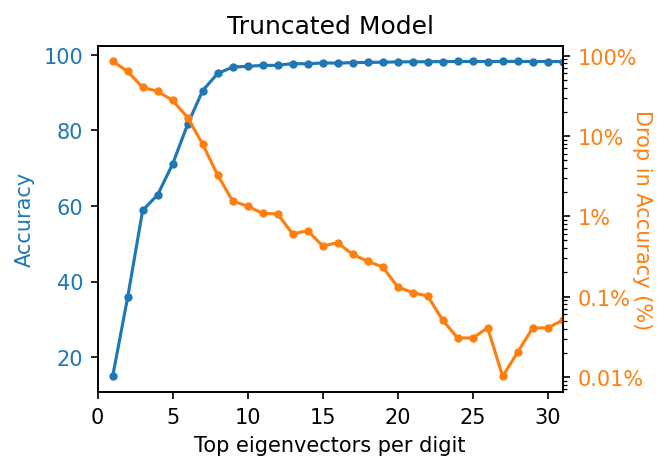}
\caption{The validation accuracy for an MNIST model truncated to the top $k$ eigenvectors by eigenvalue magnitude.}
\label{fig:accuracy}
\end{figure}

\begin{figure}[h]
\centering
\includegraphics[width=0.425\textwidth]{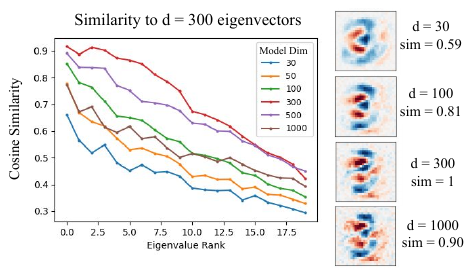}
\caption{Cosine similarities for eigenvectors that best match an eigenvector from a $d=300$ model, averaged over 5 initializations per model size. The eigenvectors for digit 3 show that moderate similarities can correspond to visually similar features. Results shown are only for positive eigenvectors and similarities are computed for eigenvectors transformed to the input basis using the embedding weights.}
\label{fig:eig_similarity}
\end{figure}

\autoref{fig:accuracy} shows the validation accuracy for an MNIST model truncated to the top-$k$ eigenvectors. Nearly full performance (less than 1\% drop in accuracy) is recovered when only 10 eigenvectors (in total, both positive and negative) per digit are kept. Generally, the percent drop in accuracy decreases exponentially with the number of eigenvectors, supporting the idea that the eigenvalues usefully parameterize the importance of a feature.

Across different model sizes and initialization, it is possible to find very similar input features. \autoref{fig:eig_similarity} looks at the cosine similarity of the best match to the positive eigenvectors in a $d=300$ model. The best matches are for models of the same size, with around 0.9 similarity for the top five or so eigenvectors per digit (roughly corresponding to a truncated model with only a 1\% drop in accuracy). Across model sizes, similar sizes have more similar eigenvectors but even a small $d=30$ model has similarities between 0.5-0.6 for the top eigenvectors. In high dimensions, such moderate cosine similarities can still correspond to visually similar features, as shown for digit 3 in \autoref{fig:eig_similarity}.

\subsection{Regularization}

Adding regularization to the model is important for getting more visually interpretable input features. We use weight decay and a regularization we call latent noise, which adds dense Gaussian noise to the inputs of each layer, including the embedding and unembedding layers. The motivation for latent noise is the finding that noise promotes the emergence of sparse feature representations in autoencoders, even when there is no L1 sparsity penalty \cite{bricken2023emergence}. We hypothesize that adding dense noise between each layer encourages a better handoff of information from one layer to the next since the next layer's features must activate strongly and robustly enough to overcome random activations from the noise. 

\begin{figure}[t]
\centering
\includegraphics[width=0.3\textwidth]{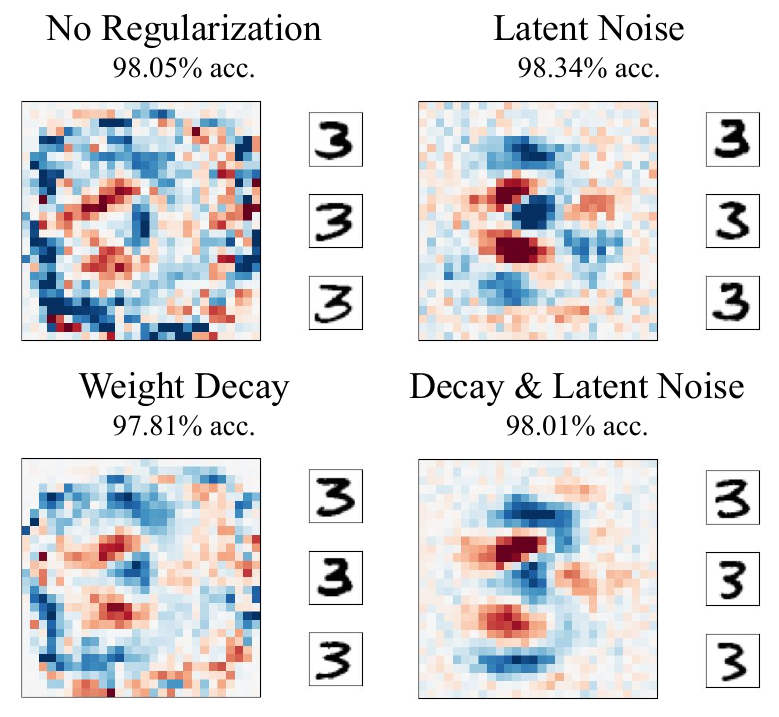}
\caption{Comparison of different regularizations. Latent noise regularization adds dense Gaussian noise to the inputs of each layer, including the embedding and unembedding layers. We use a noise with a standard deviation of 0.33 the standard deviation of the input. Latent noise is found to remove the large values in the periphery of the image and improve validation accuracy. Weight decay (parameter=0.5) reduces the fine-scale noisiness in the images and reduces performance. The paper results are from models with weight decay and latent noise.}
\label{fig:regularization}
\end{figure}

Latent noise alone improved validation accuracy from 98.05\% to 98.34\%, likely indicating better generalization. \autoref{fig:regularization} shows that the features without regularization are noisy and have large values on the periphery where pixels are rarely ``on'' which may be due to overfitting to outlier images. Training with latent noise removes the peripheral values and noisiness and adding weight decay further cleans up the features. With both types of regularization, the performance is the same as without regularization. These results suggest that regularization during training may be important for finding more interpretable weights among the large degeneracy of weights with the same performance. 

\begin{figure*}[t]
\centering
\includegraphics[width=0.9\textwidth]{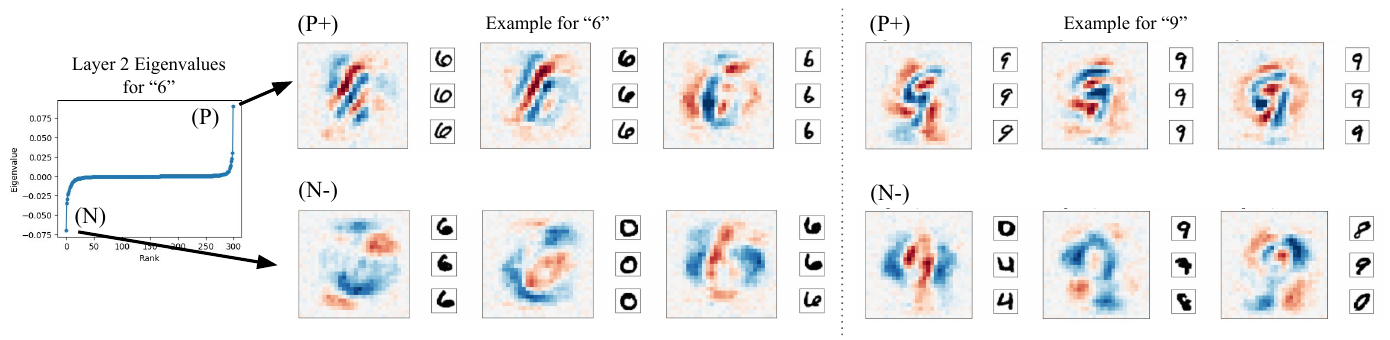}
\caption{Eigendecomposition of a two-layer model. Across digits, there is typically one prominent positive (P) and one prominent negative (N) layer-2 eigenvalue. For the examples of digits 6 and 9, we show the top 3 positive eigenvectors for (P) and the top 3 negative layer-1 eigenvectors for (N). Across digits, there is a pattern where (P+) eigenvectors have sharp lines specific to the output digit (eg, 6 or 9) while (N-) has vague shapes resembling the digit. We hypothesize that inputs for other digits that still resemble 6 or 9 will activate the (N-) eigenvectors but not the (P+) ones as strongly, leading to a negative contribution to the logits.}
\label{fig:two_layers}
\end{figure*}

\subsection{Two-Layer Eigendecomposition} 

Following the procedure outlined in \autoref{sec:full_model_eigendecomp}, we decompose a 
two-layer MNIST model. The spectra of the layer-2 eigenvalues typically have one prominent positive eigenvalue (P) and one negative one (N), as seen in the spectrum for 6 in \autoref{fig:two_layers}. The layer-1 eigenvectors for (P) and (N) have an interesting pattern. Focusing on the example of digit 6, the positive layer-1 eigenvectors (P+) have sharp lines specific to 6 while the (P-) eigenvectors have vague non-6 shapes. Likewise, (N+) has sharp non-6 lines while (N-) has vague 6 shapes. A similar pattern is shown for digit 9.

We hypothesize that this pattern, seen across the digits, is a way to better differentiate true 6s from similar non-6s. A faux-6 input will activate the vague 6-like shapes in (N-) and contribute negatively to the 6-logit. But it will be less likely to activate the more specific 6 lines of (P+). A true-6 will activate some of the (P+) eigenvectors strongly, overcoming any activation of the (N-) eigenvectors. 

\subsection{Limitations to interpretability} \label{sec:limits_to_interp}

Even though the top eigenvectors appear largely interpretable, we expect there to be potential limitations to interpretability that arise in other contexts.  
One issue is polysemanticity within eigenvectors. For example, in the spectrum for digit 1 in \autoref{fig:mnist_eigenvectors}, it is clear that the positive and negative components, particularly for the second positive eigenvector, can represent 1's with different slopes. This polysemanticity is possible due to the XOR-like behavior of the squared dot product in an activation of the eigenvector. 

More generally, sparsity in the inputs as seen for LLMs (due to vocab size and attention) will likely encourage polysemanticity in the eigenvectors as well. If the MNIST dataset contained digits that only occupied a single quadrant of the image, we would likely see eigenvector images that polysemantically contain features for each of the four quadrants. Eigendecomposition alone will not be enough to derive monosemantic features from weights. It will likely need to be combined with additional techniques, potentially dictionary learning. 

Another limitation is that eigenvectors from the same decomposition are all orthogonal. This means that certain features could be split among different eigenvectors, or an eigenvector might best be interpreted in combination with other ones that commonly co-occur. Analysis of the co-occurrence patterns across eigenvectors will help better understand the model's computations and is left to future work.

\section{Language Models} \label{sec:stories}

In this section, we present preliminary results of our techniques on a language modeling task. We use a single-layer transformer model in which the MLP is replaced with a bilinear version. 
We train this model on the Tiny Stories dataset \citep{tinystories}. We make several simplifications for ease of analysis; importantly, these modifications are not required. First, we use a simple uncased WordPiece \citep{wordpiece} tokenizer with a vocabulary size of 4096, created specifically for this dataset. Second, we omit all normalization layers and biases. The full details of our architecture and training setup are located in \autoref{app:stories_setup}.

\begin{table}[h!]
\begin{tabularx}{0.48\textwidth}{Y|l|Y}
  \textbf{Preceding} & & \textbf{Following} \\
\hline
little, the, young, small, poor, good, birthday, another, first, kind & \textbf{girl} & named, who, nodded, called, hugged, touched, names, cried, s, name \\
\end{tabularx}
\caption{The 10 highest weights corresponding to preceding and following tokens for "girl". This considers both the direct path and the path through the MLP.}
\label{fig:bigrams}
\end{table}

\begin{figure*}[!th]
\centering
\includegraphics[width=0.99\textwidth]{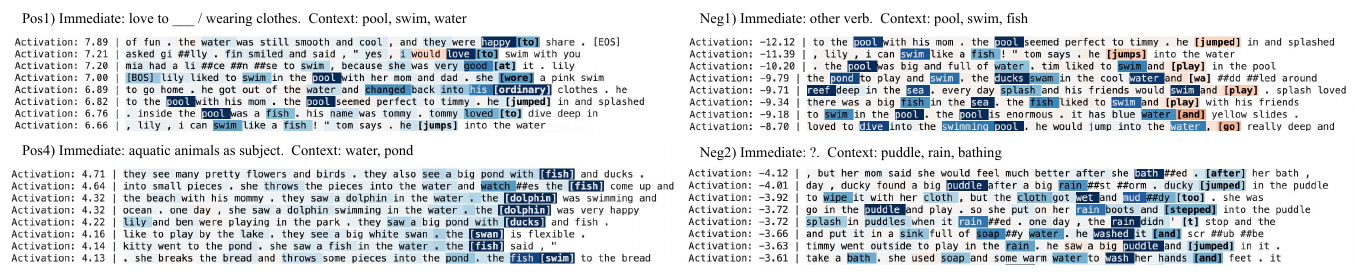}
\vspace{-0.4cm}
\caption{ Top input activations for eigenvectors of a single-layer Tiny Stories model. The output direction used is the unembedding of the ``swim'' token minus the average unembedding for other verbs [``run'', ``climb'', ``eat'', ``see'', ``smell'', ``walk'', ``fly'', ``sit'', ``sleep''], which helps remove grammatical features that apply to any verb. The coloring indicates a token's contribution to the eigenvector's activation, either positive (blue) or negative (red). The feature descriptions are divided into immediate (previous few tokens) and broader contexts. See \autoref{fig:tiny_stories_swim_minus_verbs_full} for additional eigenvectors and see \autoref{fig:tiny_stories_swim_only} for the eigenvectors for only ``swim''}
\label{fig:tiny_stories_eigs}
\end{figure*}

\subsection{Bigrams \& Skip-Trigrams}

Since the bilinear layer parameterizes interactions between pairs of inputs, it's easy to read off important interactions. For example, we can find how the MLP contributes to the learned bigrams. In ordinary models, bigrams are approximated by the direct path between the embedding and unembedding. In a bilinear layer, the direct path token through attention interacts with itself, which also contributes to the bigram. This interaction is captured in the diagonal of each interaction matrix, $B_{:ii}$, which combined with the residual path represents a more complete picture of the bigrams the model has learned. Given that no normalization is present, adding this diagonal to the residual requires no approximation. Additionally, this matrix can be transposed to show the most likely preceding tokens, as seen in \autoref{fig:bigrams}.

Generally, the embeddings learn the clearest bigrams while the MLP focuses on disambiguating bigrams related to context. For instance, it strongly impacts the token "pre", which can have multiple sensible continuations (``\#\#cisely", ``\#\#pare", ``\#\#fer") along with other compound tokens as shown in \autoref{table:bigrams}. Furthermore, it is also strongly involved in punctuation, which can also be seen as a form of context-aware bigrams. 
We find that the MLP diagonal can sometimes contain strange bigrams with high values, possibly due to noise or the inherent low rank of the interaction matrices. We defer a deeper study of this phenomenon to further work.


In a single-layer model, tokens passing through attention followed by the unembeddings will capture skip-trigrams of the form ``A...BC", as described in \citet{elhage_mathematical_2021}. In bilinear models, we can find additional contributions to skip-trigrams from the MLP by analyzing the interaction of a previous token ``A'' that passes through attention, and the direct path ``B''. The ``A'' token will be transformed by the $OV$ matrix for the specific head it passes through. For a specific output token ``C'', we can find the top interactions between ``A'' and ``B'' in the interaction matrix. 

We show the effectiveness of this approach by uncovering skip trigrams ending in a quote token. In this simple setup, quotes are exclusively predicted to open or close reported speech. This reveals that 99 out of the top 100 weights in head 0.2 correspond to skip trigrams of the form talk-related verb (say, shout, whisper) followed by a comma or a colon. The closing of the quote is performed by several heads, most notably by head 0.5, which fires on trigrams of the form quote, punctuation. We show comprehensive tables in the appendix (\autoref{app:bigram-trigram})

\subsection{Eigenvectors}

In a single-layer model, the MLP goes beyond skip trigrams by capturing interactions between the tokens that pass through the attention layer. The interactions help pick up on higher n-gram statistics, for example, the broader context of a given token. With a bilinear layer, we can read off specific token interactions, but it's difficult to describe its overall behavior. The eigenvector decomposition described in \autoref{sec:single_eigendecomp} can be a starting place for a more general understanding.

A preliminary investigation of the eigenvectors for the single-layer Tiny Stories model shows that the eigenvectors generally capture relevant aspects for predicting a given output token. \autoref{fig:tiny_stories_eigs} shows the top input activations for the eigenvectors for the output ``swim'' and each token's contributions to the activation. Initially, we found that the eigenvectors picked up on grammatical features that could apply to any verb, shown in \autoref{fig:tiny_stories_swim_only}, so we instead show results for the unembedding of ``swim'' minus the average unembedding of some other verbs. 

The eigenvectors' top activations appear mostly monosemantic. For example, having aquatic animals as subjects (Pos4) or non-swimming contexts for water such as rain, puddles, and baths (Neg2). There is a strong negative eigenvector (Neg1) that inhibits an output of ``swim'' based on the context of ``swim''-related words. In fact, this pattern is common across different output tokens. We hypothesize that the strong negative eigenvector cancels out skip-trigram (or perhaps skip-bigram) predictions already in the residual stream, in favor of higher-order predictions from the MLP. We find similar results for the eigenvectors for outputting quotation marks, shown in \autoref{fig:tiny_stories_quotation_full}.

The top positive eigenvector for ``swim'' (Pos1) is somewhat polysemantic, since it activates for the immediate context of ``love to \rule[0pt]{0.5cm}{0.4pt}" which is appropriate for verbs and the context of putting on clothes, appropriate for ``swim" as the start of ``swimsuit". Both rely on a broader context of being at the pool or near water. 

\begin{figure}[!t]
\centering
\includegraphics[width=0.29\textwidth]{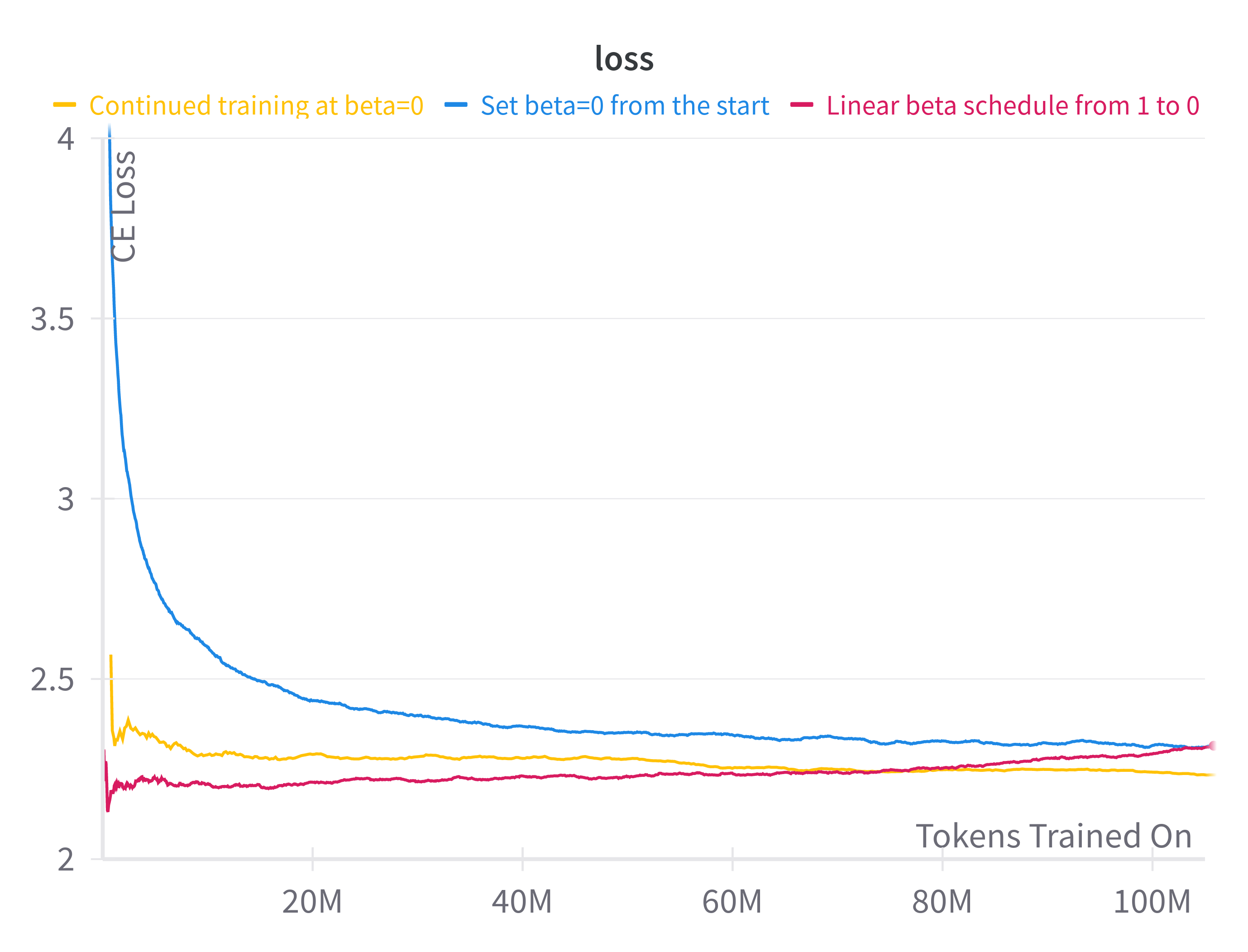} 
\caption{The result of two training runs, both trained on approximately 105 million tokens of FineWeb data. Blue: setting \(\beta\) to 0 from the start. Red: Linearly scheduling \(\beta\) over the course of the same training. Both runs converge to the same loss. Yellow: Continued training from a $\beta=0$ checkpoint.}
\label{fig:wandb_training}
\end{figure}

\subsection{Finetuning} \label{sub:finetuning}

An interpretable architecture can only see widespread adoption if it is both comparably performant and efficient to train.
In this section, we provide evidence that bilinear transformers may satisfy both conditions, as pretrained models can be finetuned into bilinear variants while using significantly less compute compared to pretraining.

Many state-of-the-art open-source models use a gated SiLU activation.\footnote{LLaMA, Mistral, Mixtral, PaLM all use SiLU. Gemma uses a GELU, which can be approximated by SwiGLU with $\beta=1.702$.} Since $\text{Swish}_\beta$ interpolates between common activation functions including SiLU ($\beta=1$) and linear ($\beta=0$) we can use a gated Swish, as in \autoref{eq:glu}, to finetune between the gated SiLU and bilinear activations.

We finetuned TinyLlama-1.1B, a 1.1 billion parameter transformer model pretrained on 3 trillion tokens of data, using a single A40 GPU. We finetuned the model in two ways. The first approach is to set $\beta$ to 0 at the start and perform standard finetuning. For the second approach, we implemented a linear scheduler to the $\beta$ parameter, gradually decreasing it from 1 to 0 over the course of training. 

\autoref{fig:wandb_training} shows the results of both training strategies. Throughout training on the same number of tokens, both runs converge to the same end loss. In yellow, we see that we benefit from continued training. More details on training can be found in \autoref{app:swiglu_tuning}. 




\section{Discussion} \label{sec:discussion}

We have presented a novel method to decompose bilinear MLPs into sets of eigenvectors by leveraging their close-to-linear structure. In the case of MNIST digit classification the top eigenvectors appear largely interpretable and help reveal interesting structure in a two-layer model. Preliminary results for a Tiny Stories language model suggest the top eigenvectors may be interpretable also. 

The method has several attractive properties for mechanistic interpretability: 
\begin{enumerate}
    \item The eigenvectors are derived entirely from the model weights and are fully equivalent to the model's original computations. So there are no concerns about data dependence or unexplained error terms, as is currently the case for sparse autoencoders.
    \item The eigenvector basis effectively sparsifies interactions since an eigenvector only interacts with itself. 
    \item The eigenvalues naturally parameterize each eigenvector's importance and show that MNIST models with small eigenvalues truncated retain good performance.
\end{enumerate}

Our method, however, requires an architecture change since it only works for bilinear MLPs. The change may be minor though, since \citet{shazeer2020glu} showed that transformers with bilinear activations perform on par with other activations: slightly worse than SwiGLU but better than ReLU. We show that LLMs can be finetuned from a SiLU to bilinear and achieve the expected loss for a pretrained bilinear model. This opens up the possibility of applying our methods to existing models by first finetuning them to have bilinear activations. 

Along the lines of \citet{bricken2023emergence}, we found that adding latent noise---dense Gaussian noise between each layer---during training helped improve the interpretability of the eigenvectors and in some cases improved model performance. Regularization may be an under-explored avenue for improving feature interpretability. 

A limitation of the eigenvector decomposition is that the eigenvectors are not expected to be fully monosemantic, as discussed in \autoref{sec:limits_to_interp}. We see evidence that MNIST eigenvectors contain different orientations of a digit in their positive and negative components. The eigenvector decomposition may be a good starting place for applying other interpretability methods, such as sparse autoencoders (SAEs). Since eigenvectors have sparse interactions, combining them with SAEs may effectively produce a transcoder with sparse interactions between sparse features, similar to those used in \citet{dunefsky2024transcoders}. Explorations in this direction are left to future work.

For a deep MLP-only bilinear model, we show it's possible to fully decompose the model into a decompiled state where interactions have been sparsified and made explicit through the tree-like computational graph. However, the number of layer-1 eigenvectors grows exponentially as $(\text{d\_model})^\text{n\_layers}$. We think this decompiled model may be useful for understanding feature construction in toy model settings but likely won't scale to large models without improvements in the method.

\section*{Acknowledgements}

We are grateful to Lee Sharkey whose paper on bilinear models inspired this work and who offered helpful comments on an early version of this work. We would also like to thank Jos\'e Oramas, Narmeen Oozeer and Nora Belrose for useful feedback on the draft. We are grateful to the \href{https://aisafety.camp/}{AI Safety Camp} program where this work first started. We thank CoreWeave for providing compute for the finetuning experiments.

\bibliographystyle{icml2024}
\bibliography{refs}

\newpage
\appendix
\onecolumn





\section{MNIST Setup} \label{app:mnist_setup}

\begin{table}[H]
    \centering
    \begin{tabular}{lr}
        \toprule
        \multicolumn{2}{c}{\textbf{Architecture}} \\
        \midrule
        \textbf{d\_model} & 300 \\
        \textbf{d\_hidden} & 300 \\
        \textbf{n\_layer} & 1-2 \\
        \midrule
        \multicolumn{2}{c}{\textbf{Training Parameters}} \\
        \midrule
        \textbf{dropout} & 0.0 \\
        \textbf{weight decay} & 0.5 (MNIST), 1.0 (F-MNIST)\\
        \textbf{latent noise} & 0.33 \\
        \textbf{batch size} & 100 \\
        \textbf{learning rate} & 0.001 \\
        \textbf{optimizer} & AdamW \\
        \textbf{schedule} & exponential decay \\
        \textbf{epochs} & 20-100\\
        \bottomrule
    \end{tabular}
    \caption{Model architecture and training setup for the MNIST models, unless otherwise stated in the text.}
    \label{tab:mnist_model_parameters}
\end{table}

\textbf{Architecture.} Our model consists of a linear embedding layer, the bilinear MLP layers, and a linear unembedding layer. For ease of analysis we removed biases in the embedding and bilinear MLP layers and did not include normalization layers. Early experiments showed that these features did not improve model performance in our shallow models. 

\textbf{Training.} For training we used the MNIST dataset [\href{http://yann.lecun.com/exdb/mnist/}{site}] consisting of 60,000 handwritten digits. We did not use data augmentation to increase the dataset size. Instead, we added latent noise---dense Gaussian noise---between each layer in the model and to the input. The noise's standard deviation equals the noise's parameter times the standard deviation of the original vector's elements.  

\section{Tensor Decompositions} 

\subsection{Change of basis}\label{sec:rewrite_b_tensor}
Given a complete or over-complete set of $m$ $\mathbf{u}$-vectors, we can re-express $\mathcal{B}$ in terms of the eigenvectors, which amounts to a change of basis. To avoid multiple contributions from similar $\mathbf{u}$-vectors, we have to use the pseudo-inverse which generalizes the inverse for non-square matrices.
Taking the $\mathbf{u}$-vectors as the columns of $U$, the pseudo-inverse $U^+$ satisfies $UU^+ = I$, as long as $U$ is full rank (equal to $d$). Then 
\begin{align}
    \mathcal{B} &= \sum^m_k \mathbf{u}^+_{:k} \otimes Q_k \\
                &= \sum^m_k \sum_i^d \lambda_{\{k,i\}} \ \mathbf{u}^+_{:k} \otimes \mathbf{v}_{\{k,i\}} \otimes \mathbf{v}_{\{k,i\}}
\end{align}
where $\mathbf{u}^+_{:k}$ are the rows of $U^+$. We can then recover the interaction matrices from $Q_k = \mathbf{u}_{k:} \cdot_\text{out} \mathcal{B}$ using the fact that
$\mathbf{u}_{k:} \cdot \mathbf{u}^+_{:k'} = \delta_{kk'}$ (Kronecker delta). 

\subsection{Higher order SVD}\label{sec:tensor_decomp}
Instead of working backwards from the outputs, we can use higher-order tensor decompositions on $\mathcal{B}$ for a given layer in isolation. The simplest approach that takes advantage of the symmetry between the inputs of $\mathcal{B}$ is to reshape the tensor, flattening the two input dimensions to produce a $d_\text{out} \times d_\text{in}^2$ shape matrix, and then do a standard singular value decomposition (SVD). 
Schematically, this gives $B_{\text{out},  \text{in}\times\text{in}} = \sum_i \sigma_i \ \mathbf{r}^{(i)}_{\text{out}} \otimes Q^{(i)}_{\text{in}\times\text{in}}$, where $Q$ can still be treated as an interaction matrix and further decomposed into eigenvectors as described in \autoref{sec:single_eigendecomp}.

\section{Tiny Stories Setup} \label{app:stories_setup}

Given that the main goal of this project is to create interpretable (not capable) models. This section delves into all architectural and training tradeoffs in designing our language models for maximal interpretability. Note that all claims in this section are based on superficial ablations. Lastly, we provide anecdotal insights that may be of interest.

\subsection{Hyperparameters}

\begin{table}[H]
    \centering
    \begin{tabular}{lr}
        \toprule
        \multicolumn{2}{c}{\textbf{Architecture}} \\
        \midrule
        \textbf{d\_model} & 1024 \\
        \textbf{d\_hidden} & 3072 \\
        \textbf{d\_head} & 128 \\
        \textbf{n\_head} & 8 \\
        \textbf{n\_layer} & 1 \\
        \textbf{n\_ctx} & 256 \\
        \textbf{n\_vocab} & 4096 \\
        \textbf{parameters} & 22,020,096 \\
        \midrule
        \multicolumn{2}{c}{\textbf{Training Parameters}} \\
        \midrule
        \textbf{dropout} & 0.0 \\
        \textbf{weight decay} & 0.1 \\
        \textbf{batch size} & 128 \\
        \textbf{learning rate} & 0.001 \\
        \textbf{optimizer} & AdamW \\
        \textbf{schedule} & linear decay \\
        \textbf{epochs} & 5 \\
        \textbf{tokens} & $\pm$ 2B \\
        \midrule
        \multicolumn{2}{c}{\textbf{Miscellaneous}} \\
        \midrule
        \textbf{pos\_emb} & rotary \\
        \textbf{initialisation} & gpt2 \\
        \bottomrule
    \end{tabular}
    \caption{Model architecture and training setup. Omitted parameters are the HuggingFace Trainer defaults.}
    \label{tab:model_parameters}
\end{table}

\subsection{Architecture}



The field of mechanistic interpretability is converging towards the limits of circuit-based analysis. This has prompted the use of leaky abstractions, such as sparse autoencoders, which may conceal noteworthy behavior. Therefore, this paper takes a different approach and modifies the architecture to obviate approximations.

\textbf{MLP.} Ordinarily, the most complex object to interpret in a transformer model is the MLP. Within our architecture, we replace this with a similar bilinear variant. Specifically, with a bilinear layer, followed by a linear out projection: $P(W \textbf{x} \odot V \textbf{x})$. Notably, keeping all else equal, this change results in 33\% more parameters in the model. As a trade-off, the expansion factor can be set to 3 (instead of 4), resulting in only an 8\% increase. From limited experiments, we found that bilinear MLPs outperform ReLU-based versions with equal parameter count, in line with \citet{shazeer2020glu}. 

\textbf{Normalization.} Empirically, we found that using normalization in our single-layer model has no impact on validation loss. Therefore, this component is removed in all locations. This strengthens the evidence that normalization is not necessary for the model but rather acts as a training regularize. However, we wish to note that including and ignoring it did not change our interpretation results significantly.

\textbf{Biases.} There seems to be little consensus on whether biases are useful in language models. In our setup, we found this not to be the case, consequently, they are excluded.

\textbf{Width.} Increasing the model width seems to improve both the accuracy and the interpretability. 

\subsection{Dataset}

The TinyStories dataset \citep{tinystories} is great for interpretability purposes. It narrows down the scope of language modeling to its bare necessities. This allows the use of tiny models, which can coherently complete this task. Unfortunately, the dataset has several undesirable properties, we outline these and discuss our solution.

\textbf{Noise.} Many stories are littered with random symbols between words. While the story itself is coherent, this unnecessarily complicates the task. We remedy this by applying Unicode normalization, stripping all accents, and removing all remaining non-ASCII symbols.

\textbf{Vocabulary.} Stories are automatically generated by GPT-3.5 and GPT-4. Unfortunately, the instruction to use a simple vocabulary is occasionally violated. Consequently, the dataset contains a considerable amount of complex words (additionally, it also contains spelling mistakes). We did not make efforts to clean this.

\textbf{Contamination.} For unknown reasons, the TinyStories dataset has a high degree of data contamination. About 15\% of training samples are equivalent and about 30\% of the validation set is also found in the training set. We solve this by combining both training and validation, filtering out identical entries, reshuffling, and resplitting.

\subsection{Tokenizer}

Most language models use tokenizers designed to achieve a good balance between accuracy and throughput. This commonly leads to complexities in terms of interpretability; specific words can be tokenized differently depending on a preceding space, capital letter, and so on. Again, we adjust our approach to maximize interpretability. Specifically, this encompasses training and using an interpretability-first tokenizer, this is achieved as follows.

\begin{enumerate}
    \item Split all non-alphabetic characters (punctuation, numbers, etc...) into separate tokens.
    \item Split between all whitespaces (whitespace is implicit between each token).
    \item Lowercase all characters (to avoid different tokenization on capital letters).
    \item Use WordPiece instead of BPE (the former results in cleaner splits).
\end{enumerate}

We optimize the tokenizer specifically on the TinyStories dataset. We find that using a vocabulary size of 4096 results in most common words being represented as a single token.

\subsection{Capability Analysis}

Given all the modifications to the model, architecture, tokenizer, and dataset, providing a meaningful quantitative analysis of model performance compared to \citet{tinystories} is challenging. We consider the automated analysis of creativity and coherence out of the scope of this work. However, to provide insight into the capabilities of the model, we share some generated text snippets. Note that this is generated by a single-layer model. Deeper models with fewer parameters easily outperform this.

\begin{quote}
    once upon a time, there was a little girl named lily. she loved to play outside in the sunshine. one day, she saw a big, red ball in the sky. she wanted to play with it, but it was too high up. lily asked her mom, " can i play with the ball? " her mom said, " no, lily. it's too high up in the sky. " lily was sad, but she didn't give up. she went to her mom and said, " mommy, can you help me get the ball? " her mom said, " sure, lily. let's go get the ball. " they went to the store and found a big, red ball. lily was so happy and said, " thank you, mommy! " they played with the ball all day long. when it was time to go home, lily said, " mommy, i had so much fun playing with the ball. " her mom smiled and said, " i'm glad you had fun, lily. " and they went home, happy and tired from playing with the ball. the end.
\end{quote}

\begin{quote}
    once upon a time, there was a little boy who was very curious. he wanted to know what was inside, so he asked his mom. his mom said, " let's go to the park and find out! " so they went to the park. when they got there, they saw a big tree. the tree had lots of leaves and branches. the boy was so excited! he ran over to it and started to play. he was having so much fun. but then, he saw something else. it was a big, green leaf! the boy was so happy. he ran over to the tree and started to play. he had so much fun! he played with his new toy all day long. when it was time to go home, the boy was sad to leave the park. he was so happy to have a new friend. he couldn't wait to come back and play with his new toy again. he was so glad he was able to have a fun day with his new toy. the end.
\end{quote}

\section{Bilinear Finetuning} \label{app:swiglu_tuning}

\begin{table}[H]
    \centering
    \begin{tabular}{lr}
        \toprule
        \multicolumn{2}{c}{\textbf{Architecture}} \\
        \midrule
        \textbf{d\_model} & 2048 \\
        \textbf{d\_hidden} & 5632 \\
        \textbf{d\_head} & 64 \\
        \textbf{n\_heads} & 32 \\
        \textbf{n\_key\_value\_heads} & 4 \\
        \textbf{n\_layer} & 22 \\
        \textbf{n\_ctx} & 2048 \\
        \textbf{n\_vocab} & 32000 \\
        \midrule
        \multicolumn{2}{c}{\textbf{Finetuning Parameters}} \\
        \midrule
        \textbf{dropout} & 0.0 \\
        \textbf{weight decay} & 0.0 \\
        \textbf{batch size} & 8 \\
        \textbf{sequence length} & 512 \\
        \textbf{learning rate} & $3 \cdot 10^{-6}$ \\
        \textbf{optimizer} & AdamW \\
        \textbf{schedule} & None \\
        \textbf{tokens} & 105M-500M \\
        \midrule
        \multicolumn{2}{c}{\textbf{Miscellaneous}} \\
        \midrule
        \textbf{pos\_emb} & rotary \\
        \bottomrule
    \end{tabular}
    \caption{Model architecture and training setup. Omitted parameters are the HuggingFace Trainer defaults.}
    \label{tab:model_parameters}
\end{table}

\textbf{Training Details}
We used a sequence length of 512, a batch size of 8, a fixed learning rate of $3 \cdot 10^{-6}$, no gradient accumulation, the AdamW optimizer with default settings, and trained with standard cross-entropy loss. We trained on sequences of FineWeb data filtered between 384 and 512 tokens. An oversight in our work is that the base model TinyLlama-1.1B was pretrained on a mix of RefinedWeb and StarCoder data, which has a different distribution than FineWeb. We finetuned the base model on 105M tokens of FineWeb data and included the result in the table below. Note that the base model finetuning had plateaued after 105M tokens, while the bilinear tuning had not plateaued yet at 500M tokens. While we cannot expect bilinear models to match SwiGLU performance, these results suggest that the gap between the bilinear tuned models and SwiGLU base models can be closed further. We emphasize that while bilinear tuning doesn't damage performance much, it may alter knowledge and circuits in a way we cannot predict. The ideal direction moving forward would be to use and study the bilinear variant.
\begin{table}[h]
\label{tab:ppl}
\begin{center}
\vspace{-2mm}
\scalebox{1.0}{
\begin{tabular}{l|ll}
 & Train Loss & Val. Loss \\
\hline
Base Model & N/A & 2.559  \\
Bilinear Tuned (105M tokens) & 2.310 & 2.450  \\
Bilinear Tuned (500M tokens)  & 2.222 & 2.354 \\
\hline
Finetuned Base Model (105M tokens)& 2.199 & 2.294
\end{tabular}
}
\caption{Train and Validation loss for TinyLlama-1.1B models on FineWeb data.}
\end{center}
\end{table}
\newpage

\section{MNIST Eigenvectors}\label{app:mnist_eigs}

\begin{figure}[h!]
\centering
\includegraphics[width=0.5\textwidth]{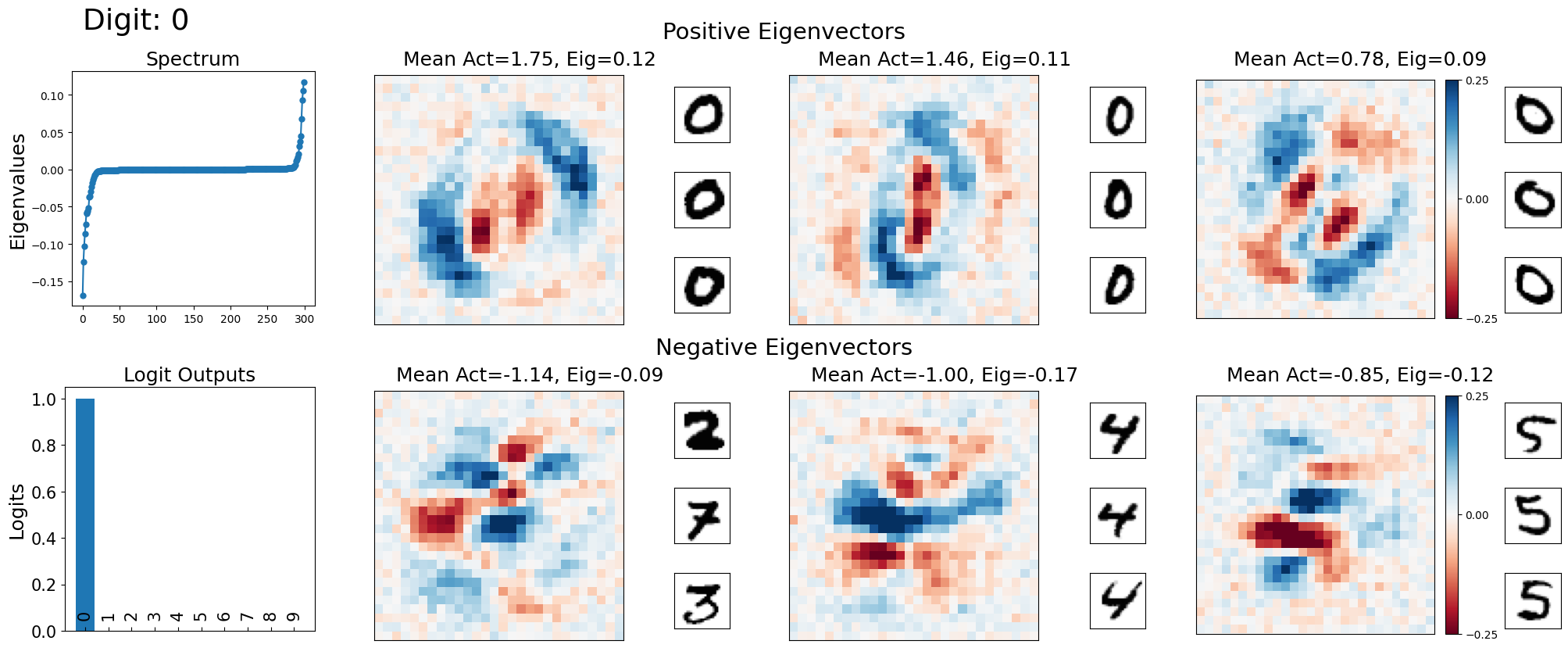}
\includegraphics[width=0.5\textwidth]{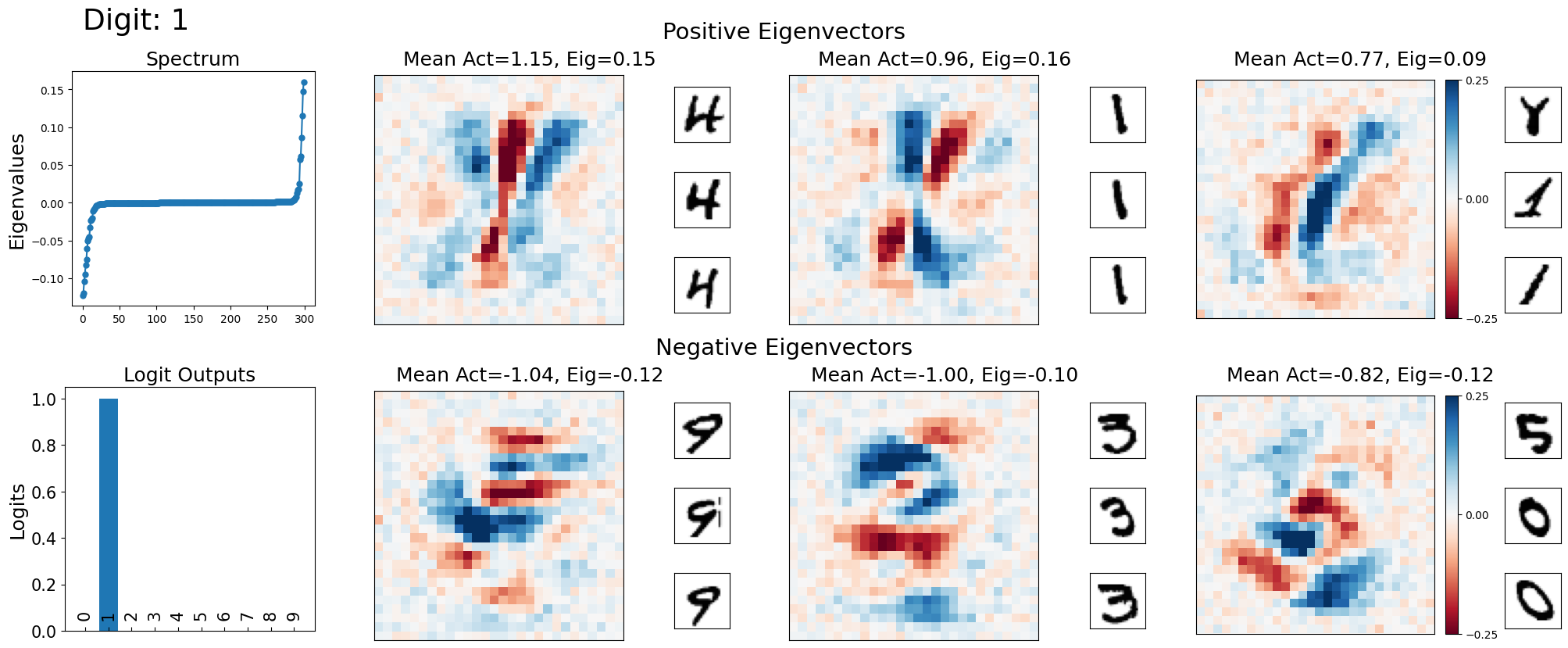}
\includegraphics[width=0.5\textwidth]{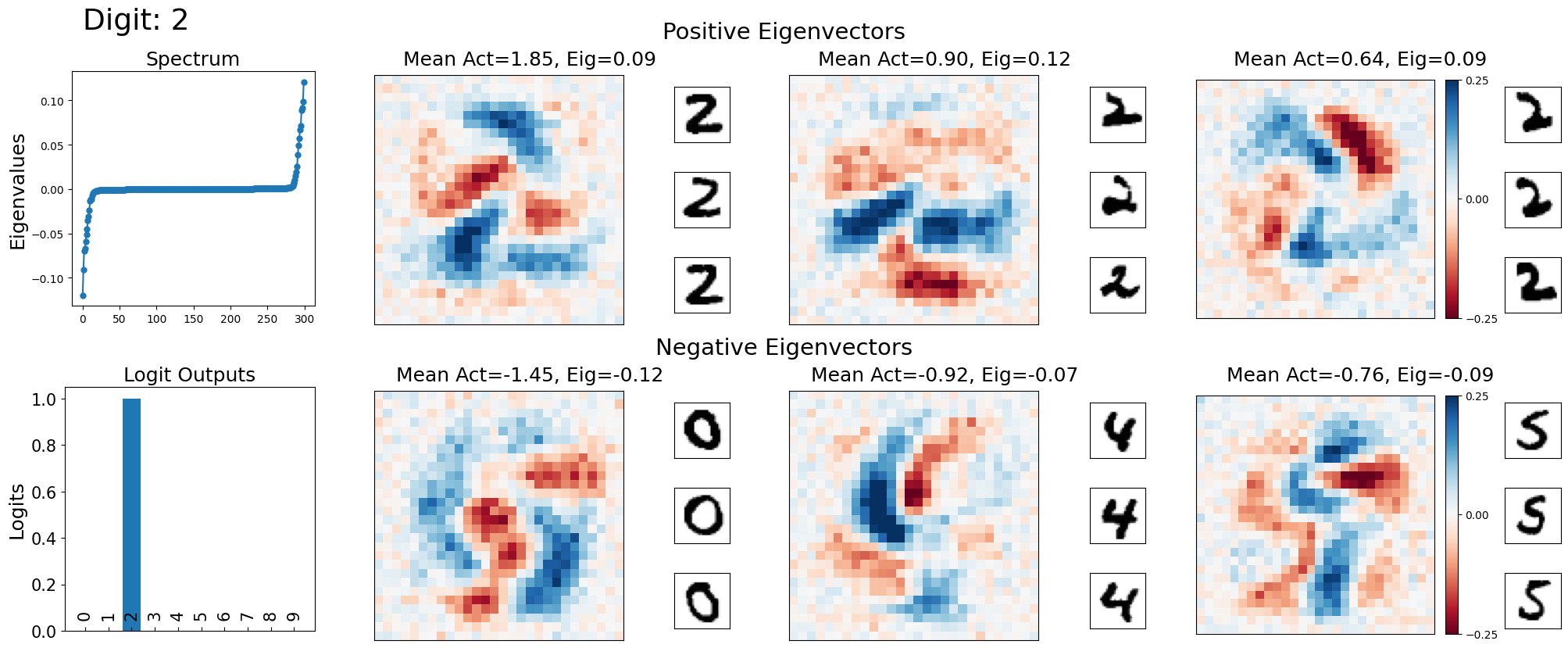}
\includegraphics[width=0.5\textwidth]{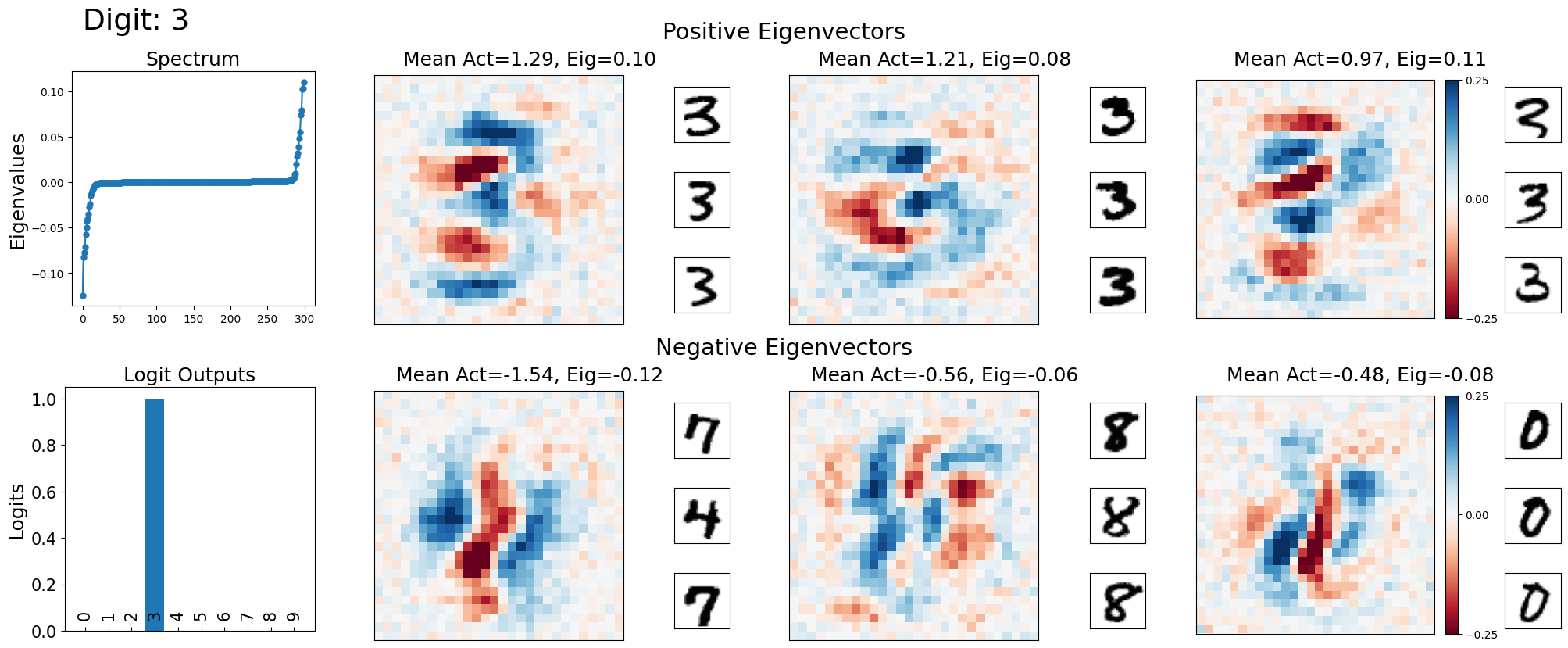}
\includegraphics[width=0.5\textwidth]{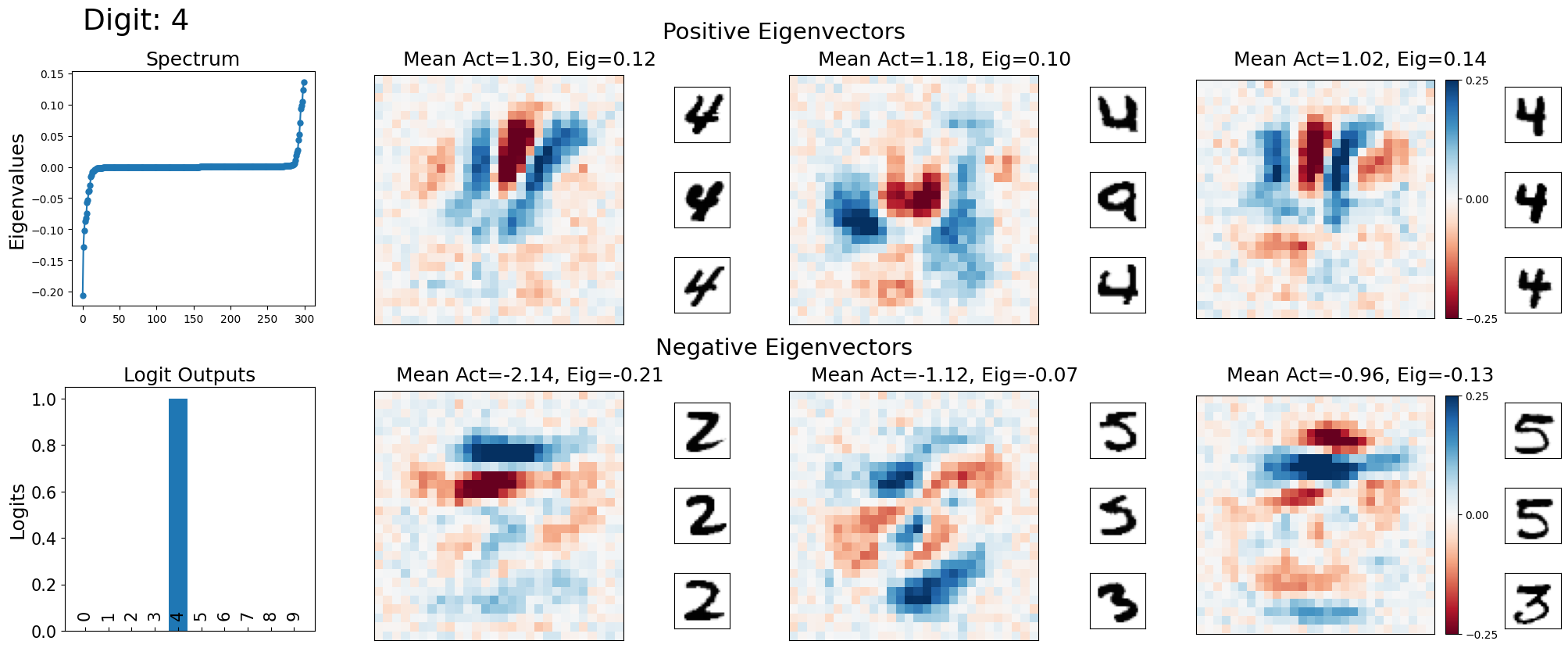}
\caption{Eigenvectors for MNIST digits 0-4}
\end{figure}

\begin{figure}[h!]
\centering
\includegraphics[width=0.5\textwidth]{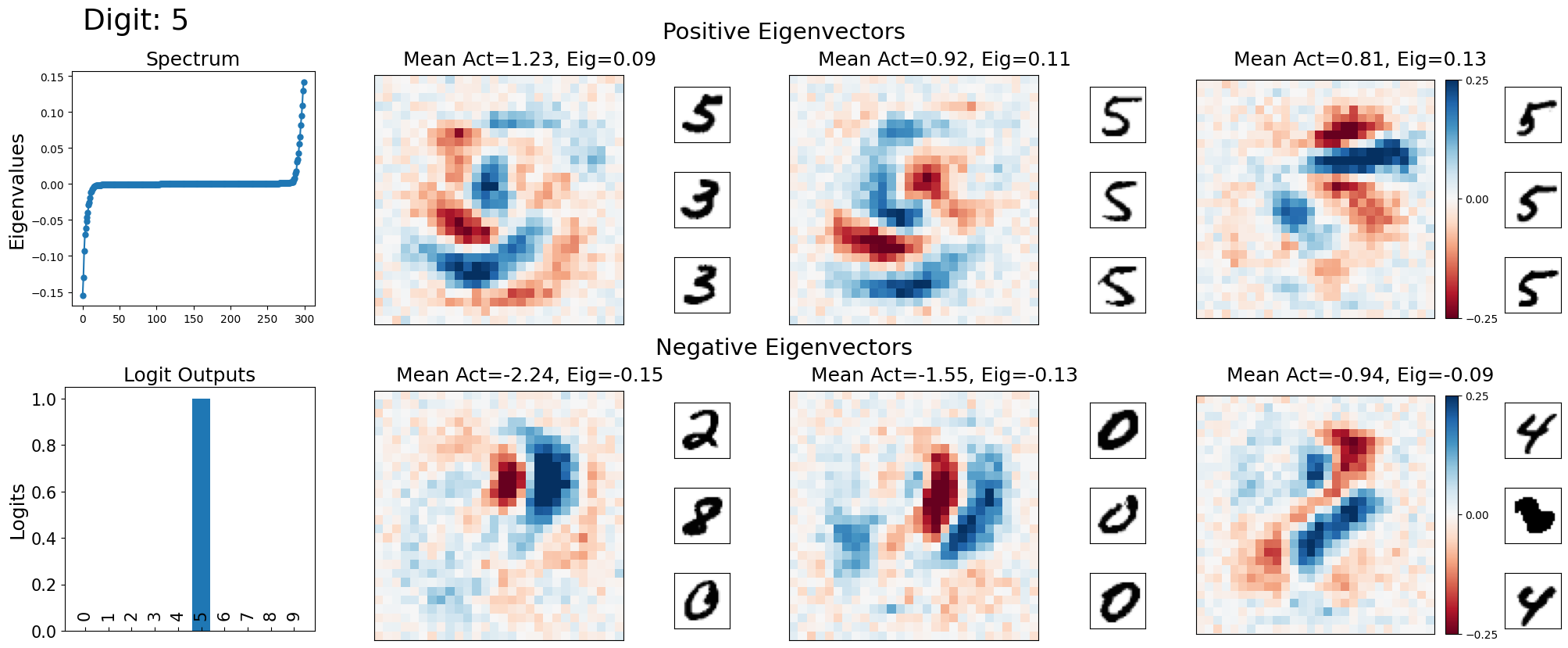}
\includegraphics[width=0.5\textwidth]{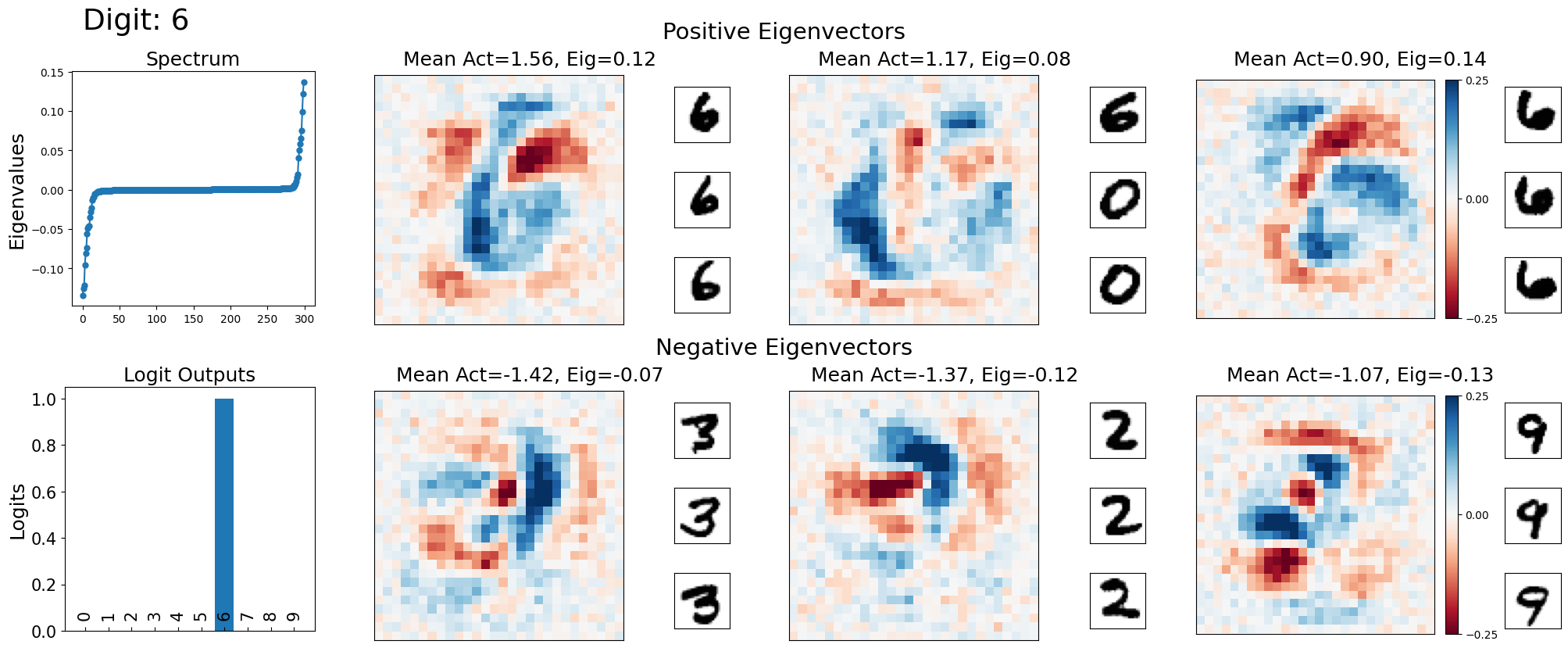}
\includegraphics[width=0.5\textwidth]{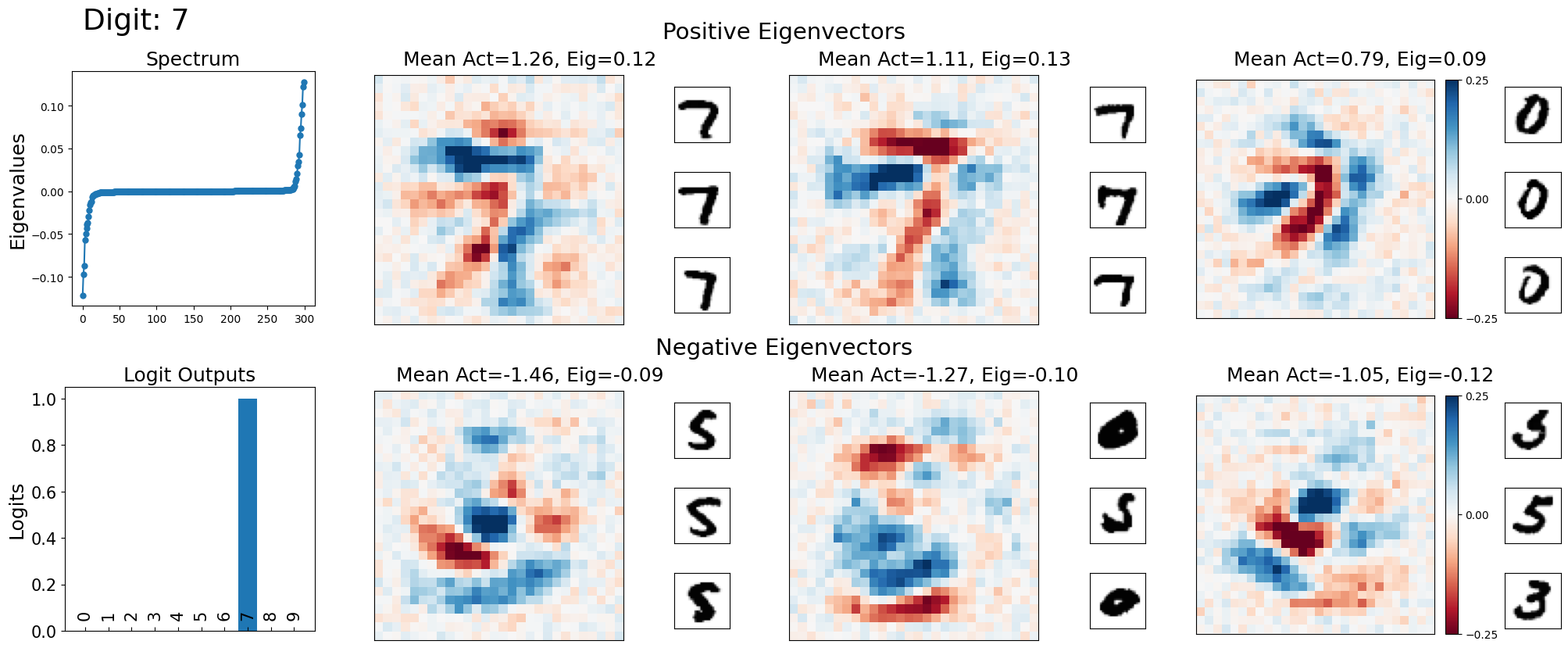}
\includegraphics[width=0.5\textwidth]{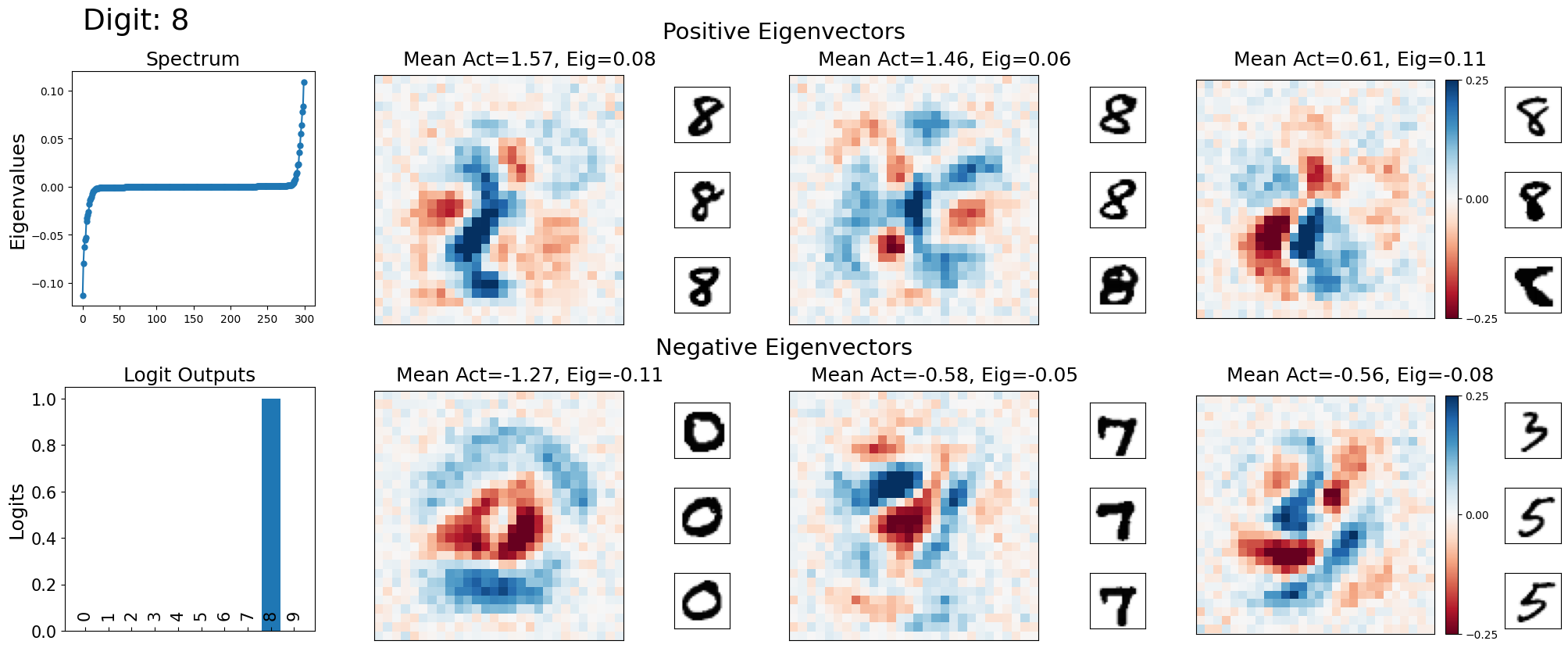}
\includegraphics[width=0.5\textwidth]{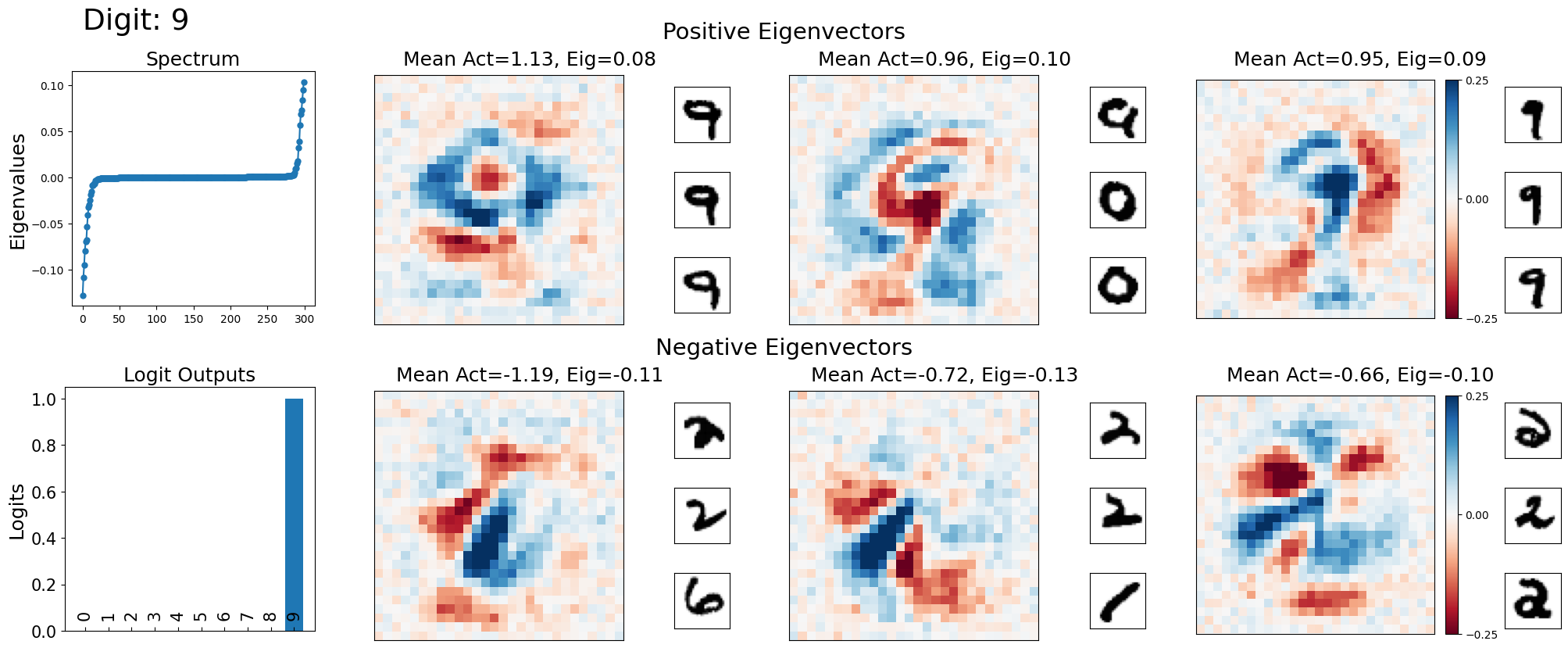}
\caption{Eigenvectors for MNIST digits 5-9}
\end{figure}
\clearpage


\section{Bigrams \& Trigrams} \label{app:bigram-trigram}

\begin{table}[h!]
    \centering
    \begin{tabular}{|l|l|l|l|l|l|}
        \hline
        \multicolumn{3}{|c|}{\textbf{Residual}} & \multicolumn{3}{c|}{\textbf{MLP}} \\ \hline
        \textbf{input} & \textbf{output} & \textbf{value} & \textbf{input} & \textbf{output} & \textbf{value} \\ \hline
        each & other & 2.41 & capt & \#\#ured & 1.14 \\ \hline
        exper & \#\#ience & 2.16 & capt & \#\#ained & 1.08 \\ \hline
        amaz & \#\#em & 2.07 & st & \#\#alk & 1.02 \\ \hline
        res & \#\#ult & 1.94 & st & \#\#ident & 0.99 \\ \hline
        re & \#\#que & 1.93 & st & \#\#ace & 0.97 \\ \hline
        e & \#\#ff & 1.90 & st & \#\#ale & 0.90 \\ \hline
        un & \#\#fort & 1.85 & pre & \#\#ci & 0.90 \\ \hline
        st & \#\#itch & 1.85 & resp & \#\#ond & 0.88 \\ \hline
        bec & \#\#om & 1.85 & pre & \#\#par & 0.85 \\ \hline
        incl & \#\#ud & 1.84 & st & \#\#ield & 0.83 \\ \hline
        thank & you & 1.82 & con & \#\#cer & 0.82 \\ \hline
        bu & \#\#dge & 1.81 & st & \#\#ained & 0.82 \\ \hline
        bran & \#\#d & 1.81 & disag & \#\#reed & 0.81 \\ \hline
        ' & s & 1.81 & gl & \#\#itter & 0.81 \\ \hline
        pl & \#\#ucked & 1.78 & des & \#\#ire & 0.78 \\ \hline
        stra & \#\#ight & 1.76 & con & \#\#ut & 0.77 \\ \hline
        purp & \#\#ose & 1.75 & bl & \#\#ouse & 0.76 \\ \hline
        fa & \#\#shion & 1.75 & bl & \#\#ush & 0.76 \\ \hline
        dir & \#\#ect & 1.75 & capt & \#\#ain & 0.76 \\ \hline
        fl & \#\#ight & 1.74 & somet & \#\#ac & 0.76 \\ \hline
    \end{tabular}
    \caption{Comparison of most salient bigrams in the weights of the Residual and MLP measured in a non-regularized model with d\_model=512. We note that the MLP generally learns bigrams that are dependent on context, such as completions of very general tokens: st, pre, ... but tends to overcompensate towards that.}
    \label{table:bigrams}
\end{table}

\begin{table}[!ht]
    \centering
    \begin{tabular}{|l|l|l|l|l|l|l|}
    \hline
        ~ & \multicolumn{3}{c|}{\textbf{Head 0.2}} & \multicolumn{3}{c|}{\textbf{Head 0.5}} \\ \hline
        ~ & virtual & direct & value & virtual & direct & value \\ \hline
        0 & saying & , & 3.79 & " & , & 18.67 \\ \hline
        1 & shouts & , & 3.76 & " & ! & 15.27 \\ \hline
        2 & say & , & 3.69 & " & : & 14.89 \\ \hline
        3 & says & , & 3.64 & " & ? & 13.18 \\ \hline
        4 & shouted & , & 3.60 & " & ; & 11.98 \\ \hline
        5 & yelled & , & 3.45 & " & . & 11.23 \\ \hline
        6 & yell & , & 3.40 & " & - & 10.58 \\ \hline
        7 & shout & , & 3.27 & " & say & 8.12 \\ \hline
        8 & \#\#laimed & , & 3.25 & " & $\backslash$ & 7.19 \\ \hline
        9 & asks & , & 3.24 & " & ' & 6.78 \\ \hline
        10 & said & , & 3.20 & " & exclaimed & 6.71 \\ \hline
        11 & \#\#laimed & : & 3.20 & " & / & 6.32 \\ \hline
        12 & said & : & 3.18 & " & when & 6.29 \\ \hline
        13 & shouted & : & 3.18 & . & , & 5.75 \\ \hline
        14 & shouting & , & 3.17 & " & says & 5.71 \\ \hline
        15 & says & : & 3.15 & " & wasn & 5.36 \\ \hline
        16 & whispered & , & 3.11 & " & shouted & 5.32 \\ \hline
        17 & saying & : & 3.08 & " & saying & 4.98 \\ \hline
        18 & asked & , & 3.04 & " & because & 4.92 \\ \hline
        19 & told & , & 3.02 & " & if & 4.88 \\ \hline
    \end{tabular}
    \caption{Strongest weights for predicting a quote for both Head 0.2 and Head 0.5. Head 0.2 detects when it should open a quote from its context. Head 0.5 is much stronger and detects when it should close a quote but also encourages the opening of a new quote if there already has been one.}
    \label{table:skip-trigram}
\end{table}

\clearpage
\section{Tiny Stories Eigenvectors} \label{app:tiny_stories_eigs}

\begin{figure*}[h]
\centering
\includegraphics[width=0.95\textwidth]{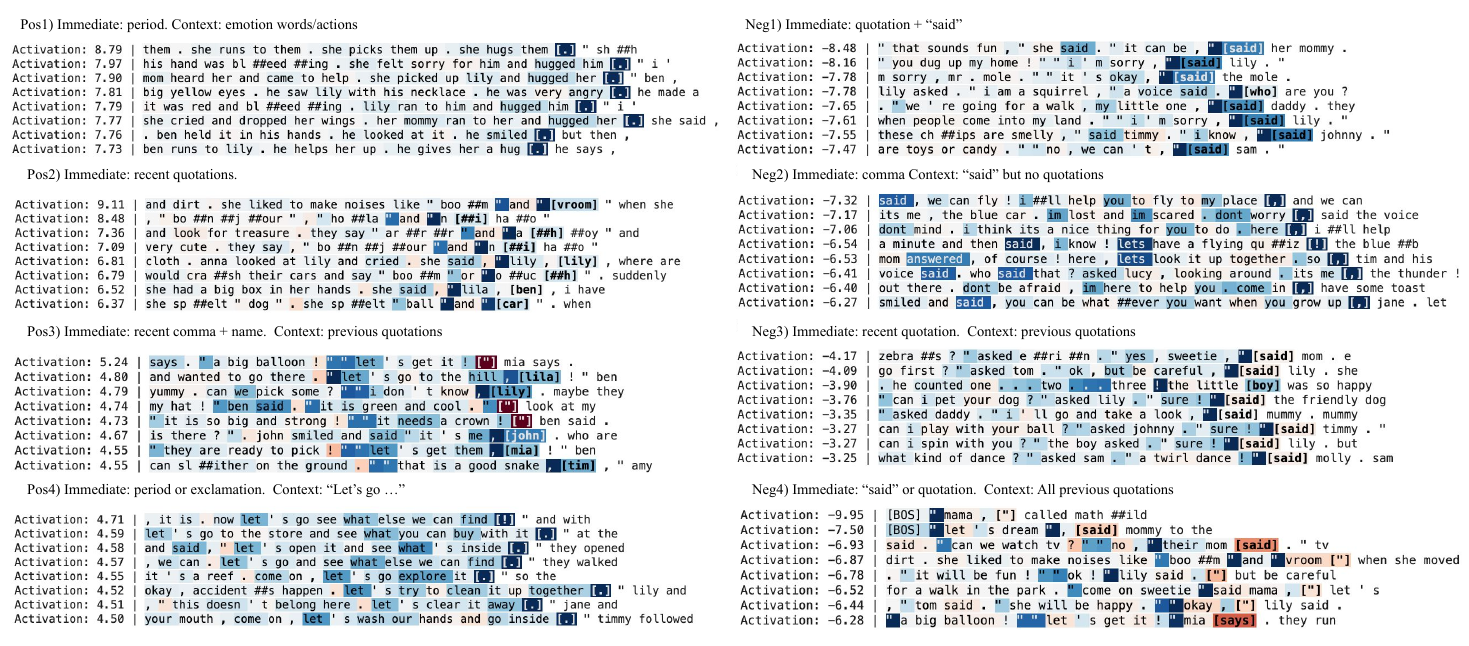}
\caption{Eigenvectors for the quotation output token (") in a single-layer Tiny Stories model.}
\label{fig:tiny_stories_quotation_full}
\end{figure*}

\begin{figure*}[h]
\centering
\includegraphics[width=0.95\textwidth]{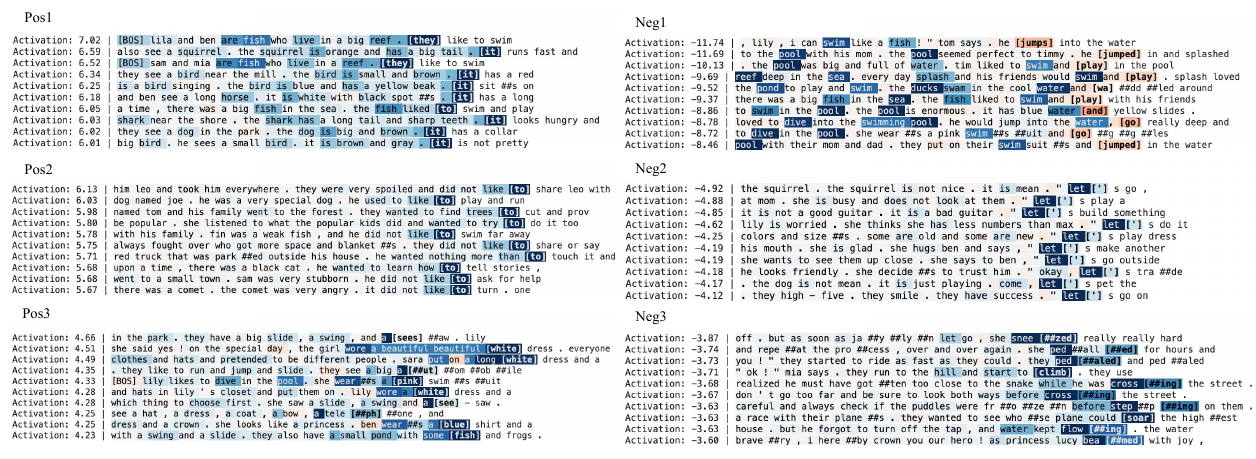}
\caption{Eigenvectors for the output token ``swim'' in a single-layer Tiny Stories model. Note that some eigenvectors capture features that grammatically apply to most verbs, such as completing the phrase ``love to \makebox[1cm]{\hrulefill}'' (Pos2) or completing the contraction ``let's`` (Neg2) where no verb would work.}
\label{fig:tiny_stories_swim_only}
\end{figure*}

\begin{figure*}[h]
\centering
\includegraphics[width=0.95\textwidth]{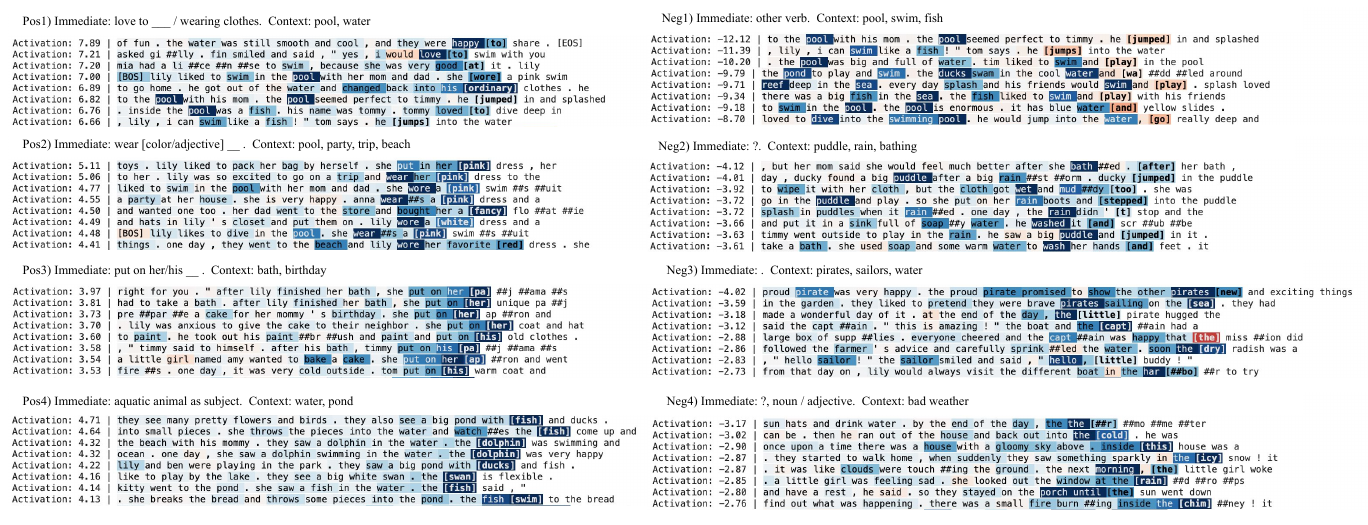}
\caption{Eigenvectors in a single-layer Tiny Stories model for the ``swim'' unembedding minus the average unembedding of [``run'', ``climb'', ``eat'', ``see'', ``smell'', ``walk'', ``fly'', ``sit'', ``sleep''].}
\label{fig:tiny_stories_swim_minus_verbs_full}
\end{figure*}


\end{document}